\documentclass[10pt,twocolumn,letterpaper]{article}
\usepackage{iccv}
\usepackage{times}
\usepackage{epsfig}
\usepackage{graphicx}
\usepackage{amsmath}
\usepackage{amssymb}
\usepackage{multirow}
\usepackage{algorithmicx,algorithm}
\usepackage{algpseudocode}
\usepackage{booktabs}
\usepackage{bbm}
\usepackage{array}
\def\ie{{\em i.e.}}
\def\eg{{\em e.g.}}

\usepackage{color}

\iccvfinalcopy 
\ificcvfinal
\fi

\usepackage[pagebackref=true,breaklinks=true,letterpaper=true,colorlinks,bookmarks=false]{hyperref}

\ificcvfinal\fi

\begin{document}
\title{Contrastive Context-Aware Learning for 3D High-Fidelity Mask Face Presentation Attack Detection}

\author{
	Ajian Liu$^{\rm 1 }$\thanks{Equal contribution} , 
	Chenxu Zhao$^{\rm 2*}$, 
	Zitong Yu$^{\rm 3*}$, 
	Jun Wan$^{\rm 4}$\thanks{Corresponding author}, 
	Anyang Su$^{\rm 2}$, 
	Xing Liu$^{\rm 2}$, 
	Zichang Tan$^{\rm 4}$ \\
	Sergio Escalera$^{\rm 5}$, 
	Junliang Xing$^{\rm 4}$, 
	Yanyan Liang$^{\rm 1}$, 
	Guodong Guo$^{\rm 6}$, 
	Zhen Lei$^{\rm 4}$, 
	Stan Z. Li$^{\rm 1,7}$, 
	Du Zhang$^{\rm 1}$ \\
	$^{\rm 1}$MUST, Macau \quad 
	$^{\rm 2}$Mininglamp Academy of Sciences, Mininglamp Technology, China \\
	$^{\rm 3}$University of Oulu, Finland \quad  
	$^{\rm 4}$NLPR, CASIA, UCAS, China \quad 
	$^{\rm 5}$CVC, UB, Spain \\
	$^{\rm 6}$Baidu Research, China \quad 
	$^{\rm 7}$Westlake University, China \\
	\tt\footnotesize
	ajianliu92@gmail.com, 
	\tt\footnotesize
	zhaochenxu@mininglamp.com, 
	\tt\footnotesize
	zitong.yu@oulu.fi \\
	\tt\footnotesize
	jun.wan@ia.ac.cn, 
	\tt\footnotesize
	sergio@maia.ub.es
}

\maketitle

\begin{abstract}
	Face presentation attack detection (PAD) is essential to secure face recognition systems primarily from high-fidelity mask attacks. 
	Most existing 3D mask PAD benchmarks suffer from several drawbacks: 1) a limited number of mask identities, types of sensors, and a total number of videos; 2) low-fidelity quality of facial masks. 
	Basic deep models and remote photoplethysmography (rPPG) methods achieved acceptable performance on these benchmarks but still far from the needs of practical scenarios. To bridge the gap to real-world applications, we introduce a large-scale \textbf{Hi}gh-\textbf{Fi}delity \textbf{Mask} dataset, namely \textbf{CASIA-SURF HiFiMask} (briefly HiFiMask). Specifically, a total amount of $54,600$ videos are recorded from $75$ subjects with $225$ realistic masks by $7$ new kinds of sensors. 
	Together with the dataset, we propose a novel \textbf{C}ontrastive \textbf{C}ontext-aware \textbf{L}earning framework, namely \textbf{CCL}. CCL is a new training methodology for supervised PAD tasks, which is able to learn by leveraging rich contexts accurately (e.g., subjects, mask material and lighting) among pairs of live faces and high-fidelity mask attacks. Extensive experimental evaluations on HiFiMask and three additional 3D mask datasets demonstrate the effectiveness of our method. 
	\footnote{We will release the code and dataset when this paper is accepted.}
\end{abstract}

\section{Introduction}
Face presentation attack detection (PAD) aims to secure a face recognition system from malicious presentation attacks (PAs), such as print attacks~\cite{Zhang2012A}, video replay attacks~\cite{Chingovska_BIOSIG-2012}, and 3D Mask attacks~\cite{ERDOGMUS_BTAS-2013}. In recent years, face PAD approaches~\cite{qin2019learning,yu2020face,zhang2020face,liu2020disentangling} for 2D attacks have made great progress, benefiting from the release of several large-scale, high-quality benchmark datasets~\cite{Boulkenafet2017OULU,Liu2018Learning,zhang2019dataset,liu2021casia}. However, with the maturity of 3D printing technology, face mask has become a new type of PA to threaten face recognition systems' security. Compared with traditional 2D PAs, face masks are more realistic in terms of color, texture, and geometry structure, making it easy to fool a face PAD system designed based on coarse texture~\cite{yang2014learn} and facial depth information~\cite{Liu2018Learning}. Fortunately, some works have been devoted to 3D mask attacks, including design of datasets~\cite{nesli2013spoofing,liu20163d,li2018unsupervised,george2019biometric,yu2020nasfas} and algorithms~\cite{kose2014mask,steiner2016reliable,liu20163d,li2016generalized,liu2018remote,lin2019face}.
\begin{figure}
	\vspace{-0.6em}
	\centering
	\includegraphics[scale=0.45]{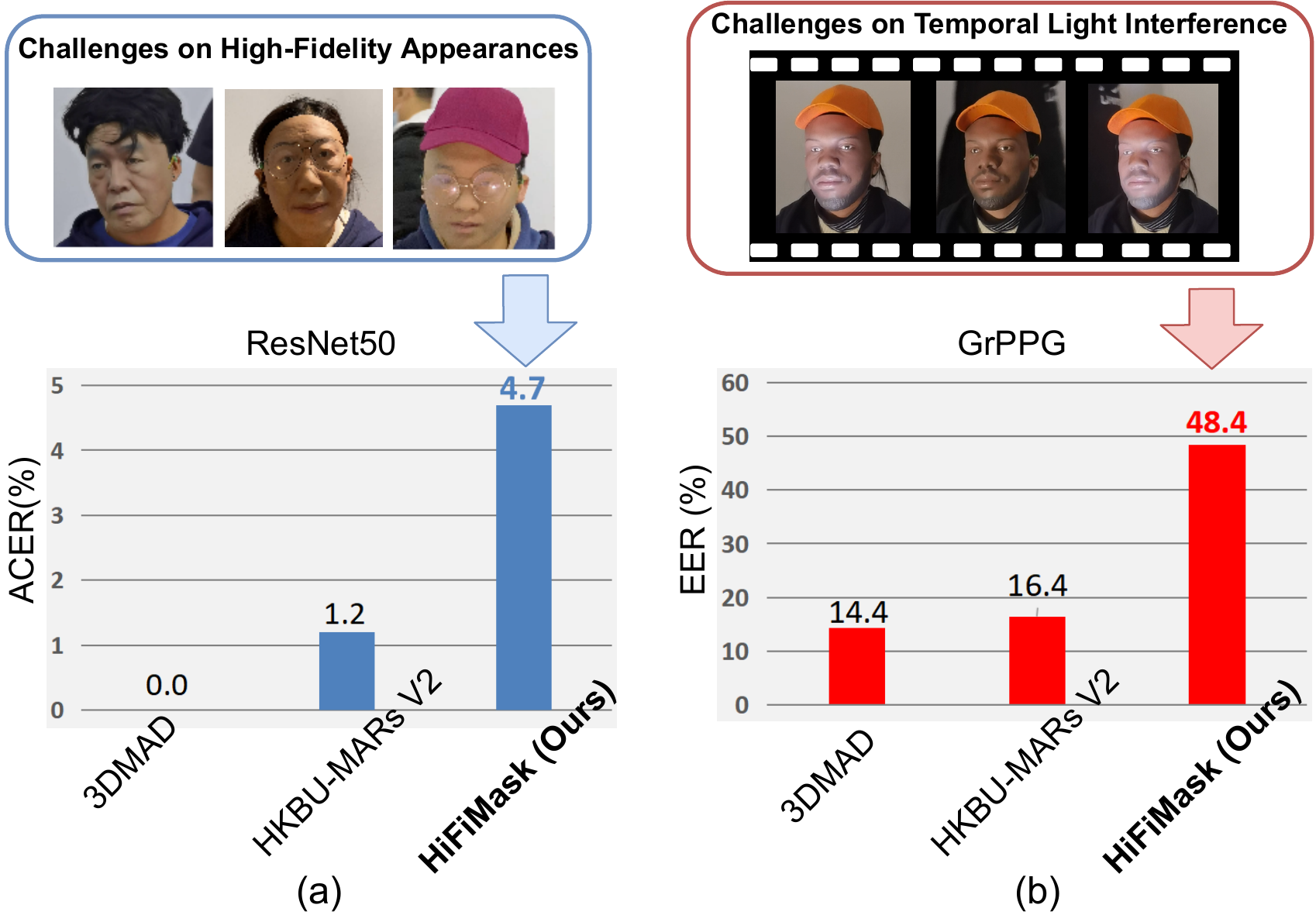}
	\caption{\small{
			Performance of ResNet50~\cite{He_2016_CVPR} and GrPPG~\cite{li2016generalized} on 3DMAD, MARsV2 and our proposed HiFiMask datasets. Despite satisfying mask PAD performance on 3DMAD and MARsV2, these methods fail to achieve convincing results on HiFiMask.
	}}
	\label{fig:Figure1}
	\vspace{-0.6em}
\end{figure}

\begin{table*}[]
	\renewcommand\arraystretch{0.8}
	\centering
	\setlength{\tabcolsep}{4pt}
	\caption{Comparison of the public 3D face anti-spoofing datasets. Y, W, and B are shorthand for yellow, white, and black, respectively. `Sub.', `Mask Id.' and `Light. Cond.' denote `Live subjects', `Mask identity numbers' and `Lighting Condition', respectively. Number with `*' denotes this number is statistically inferred and there may be inaccuracies.}
	\footnotesize{\scalebox{0.85}{
			\begin{tabular}{|l|l|l|l|l|l|l|l|l|}
				\hline
				Dataset, Year   & Skin tone & \#Sub. & \#Mask Id. & Material    & Scenes & Light. Cond.  & Devices  
				& \begin{tabular}[c]{@{}l@{}}\#Videos\\ (\#Live/\#Fake)\end{tabular} \\ \hline
				
				3DMAD~\cite{nesli2013spoofing}, 2013 & W/B       & 17          & 17         
				& \begin{tabular}[c]{@{}l@{}}Paper, hard resin\end{tabular} & Controlled     & Adjustment   & Kinect 
				& \begin{tabular}[c]{@{}l@{}}255(170/85)\end{tabular}            \\ \hline
				
				3DFS-DB~\cite{galbally2016three}, 2016  & Y         & 26          & 26         & Plastic    & Office    & Adjustment
				& \begin{tabular}[c]{@{}l@{}}Kinect, Carmine 1.09\end{tabular}  
				& \begin{tabular}[c]{@{}l@{}}520(260/260)\end{tabular}            \\ \hline
				
				\begin{tabular}[c]{@{}l@{}}BRSU~\cite{steiner2016reliable},  2016\end{tabular} & Y/W/B     & 137         & 6 
				& \begin{tabular}[c]{@{}l@{}}Silicone, Plastic\\ Resin, Latex\end{tabular} 
				& \begin{tabular}[c]{@{}l@{}}Disguise\\ Counterfeiting\end{tabular} & Adjustment.  
				& \begin{tabular}[c]{@{}l@{}}SWIR, Color\end{tabular}    
				& \begin{tabular}[c]{@{}l@{}}141(0/141)\end{tabular} \\ \hline
				
				\begin{tabular}[c]{@{}l@{}}MARsV2~\cite{liu20163d2}, 2016\end{tabular}                   & Y         & 12          & 12         
				& \begin{tabular}[c]{@{}l@{}}ThatsMyFace\\ REAL-F\end{tabular}  & Office 
				& \begin{tabular}[c]{@{}l@{}}Room light, Low light\\ Bright light, Warm light\\ Side light, Up side light\end{tabular} 
				& \begin{tabular}[c]{@{}l@{}}Logitech C920, Industrial Cam\\ EOS M3, Nexus 5, iPhone 6 \\ Samsung S7, Sony Tablet S\end{tabular} 
				& \begin{tabular}[c]{@{}l@{}}1008 \\ (504/504)\end{tabular} \\ \hline
				
				SMAD~\cite{manjani2017detecting}, 2017  & -         & Online      & Online     & Silicone & -     & varying lighting & Varying cam.
				& \begin{tabular}[c]{@{}l@{}}130(65/65)\end{tabular}             \\ \hline
				
				MLFP~\cite{agarwal2017face}, 2017  & W/B       & 10          & 7          
				& \begin{tabular}[c]{@{}l@{}}Latex, Paper\end{tabular}                      
				& \begin{tabular}[c]{@{}l@{}}Indoor, Outdoor\end{tabular}   & Daylight  
				& \begin{tabular}[c]{@{}l@{}}Visible \\ Near infrared, Thermal\end{tabular}                                                     
				& \begin{tabular}[c]{@{}l@{}}1350 \\ (150/1200)\end{tabular}          \\ \hline
				
				ERPA~\cite{bhattacharjee2017you}, 2017  & W/B       & 5           & 6          
				& \begin{tabular}[c]{@{}l@{}}Resin, Silicone\end{tabular} & Indoor & Room light  
				& \begin{tabular}[c]{@{}l@{}}Xenics Gobi, thermal cam.\\ Intel Realsense SR300\end{tabular} & 86       \\ \hline
				
				WMCA~\cite{george2019biometric}, 2019   & Y/W/B     & 72          & 7*         
				& \begin{tabular}[c]{@{}l@{}}Plastic\\ Silicone, Paper\end{tabular}         & Indoor        
				& \begin{tabular}[c]{@{}l@{}}Office light\\ LED lamps, day-light\end{tabular}   
				& \begin{tabular}[c]{@{}l@{}}Intel RealSense SR 300\\ Seek Thermal, Compact PRO.\end{tabular}                                 
				& \begin{tabular}[c]{@{}l@{}}1679 \\ (347/1332)\end{tabular}          \\ \hline 
				
				\begin{tabular}[c]{@{}l@{}}CASIA-SURF \\ 3DMask~\cite{yu2020nasfas}, 2020\end{tabular}   & Y & 48 & 48  & Plaster 
				& \begin{tabular}[c]{@{}l@{}}Indoor, Outdoor\end{tabular}                                      
				& \begin{tabular}[c]{@{}l@{}}Normal light, Back light\\ Front light, Side light\\ Sunlight, Shadow\end{tabular}        
				& \begin{tabular}[c]{@{}l@{}}Apple, Huawei\\ Samsung\end{tabular}                                                                  
				& \begin{tabular}[c]{@{}l@{}}1152 \\ (288/864)\end{tabular}           \\ \hline
				
				HiFiMask (ours), 2021   & Y/W/B     & 75          & 75         
				& \begin{tabular}[c]{@{}l@{}}Transparent \\ Plaster, Resin\end{tabular}     
				& \begin{tabular}[c]{@{}l@{}}White, Green\\ Tricolor, Sunshine\\ Shadow, Motion\end{tabular} 
				& \begin{tabular}[c]{@{}l@{}}NormalLight, DimLight\\ BrightLight, BackLight\\ SideLight, TopLight\end{tabular}      
				& \begin{tabular}[c]{@{}l@{}}iPhone11, iPhoneX\\ MI10, P40, S20\\ Vivo, HJIM\end{tabular}                                     
				& \begin{tabular}[c]{@{}l@{}}54,600\\ (13,650/40,950)\end{tabular}   \\ \hline
			\end{tabular}
	}}
	\label{table:datasets}
	\vspace{-1.6em}
\end{table*}

In terms of the composition of 3D mask datasets, several drawbacks limit the generalization ability of data-driven algorithms. From existing 3D mask datasets shown in Tab.~\ref{table:datasets}, one can see some of these drawbacks: (1) \textbf{Overfitting Problems}. The number of mask subjects is less than the number of real face subjects. Even for some public datasets as~\cite{steiner2016reliable,manjani2017detecting,agarwal2017face,george2019biometric}, the mask and live subjects correspond to completely different identities, which may produce the model to mistake identity as a discriminative PAD-related feature; (2) \textbf{Limited number of subjects and low skin tone variability}. Most datasets contain less than $50$ subjects, with low or unspecified skin tone variability. Skin tone has been verified to severely interfere with the PA detection performance~\cite{liu2021casia}; (3) \textbf{Limited diversity of mask materials}. Most datasets~\cite{nesli2013spoofing,galbally2016three,liu20163d2,manjani2017detecting,agarwal2017face,bhattacharjee2017you,yu2020nasfas} provide less than $3$ mask materials, which makes it difficult to cover the attack masks that attackers may use; (4) \textbf{Few scene settings}. Most datasets~\cite{nesli2013spoofing,galbally2016three,liu20163d2,bhattacharjee2017you} only consider single deployment scenarios, without covering complex real-world scenarios; (5) \textbf{Controlled lighting environment}. Lighting changes pose a great challenge to the stability of rPPG-based PAD methods~\cite{li2016generalized}. However, all existing mask datasets avoid this by setting the lighting to a fixed value, \ie, daylight, office light; 
(6) \textbf{Obsolete acquisition devices}. 
Many datasets use outdated acquisition devices regarding the resolution and imaging quality. Due to the above shortcomings, the number of videos in the current datasets is less than $2,000$, which does not fulfill the training needs of very deep CNNs.

In order to alleviate previous issues, we introduce a large-scale 3D High-Fidelity Mask dataset for face PAD, namely \textbf{CASIA-SURF HiFiMask} (briefly HiFiMask). As shown in Tab.~\ref{table:datasets}, HiFiMask provides $25$ subjects with yellow, white, and black skin tones to facilitate fair artificial intelligence (AI) and alleviate skin-caused biases (a total of $75$ subjects). Each subject provides $3$ kinds of high-fidelity masks with different materials (\ie, plaster, resin, and transparent). Thus, a total of $225$ masks are collected. In terms of recording scenarios, we consider $6$ scenes, including indoor and outdoor environments with extra $6$ directional and periodic lighting. As for the sensors for video recording, $7$ mainstream imaging devices are used. In total, we collected $54,600$ videos, of which the live and mask videos are $13,650$ and $40,950$, respectively. 

For 3D face PAD approaches, both appearance-based~\cite{yu2020nasfas,jia2020survey,jia20203d,george2019biometric} and remote photoplethysmography (rPPG)-based~\cite{li2016generalized,liu20163d,liu2018remote,lin2019face,liu2021multi} methods have been developed. 
As illustrated in Fig.~\ref{fig:Figure1}, although both appearance-based method ResNet50~\cite{He_2016_CVPR} and rPPG-based method GrPPG~\cite{li2016generalized} perform well on 3DMAD~\cite{nesli2013spoofing} and HKBU-MARs V2 (briefly named MARsV2)~\cite{liu20163d2} datasets, these methods fail to achieve high performance on the proposed HiFiMask dataset. On the one hand, the high-fidelity appearance of 3D masks makes it harder to be distinguished from the bonafide. On the other hand, temporal light interference 
leads to pseudo `liveness' cues for even 3D masks, which might confuse the rPPG-based attack detector.

To tackle the challenges about high-fidelity appearance and temporal light interference, we propose a novel Contrastive Context-aware Learning framework, namely \textbf{CCL}, which learns discriminability by comparing image pairs with diverse contexts. Various kinds of image pairs are organized according to the context attribute types, which provide rich and meaningful contextual cues for representation learning. For instance, constructing face pairs from the same identify with both bonafide (\ie, skin material) and mask presentation (\ie, resin material) could benefit the fine-grained material features learning. 
Due to the significant appearance variations between some `hard' positive pairs, the proposed CCL framework's convergence might sometimes be unstable. To alleviate the influence of such `outlier' pairs and accelerate convergence, the Context Guided Dropout module, namely \textbf{CGD}, is proposed for robust contrastive learning via adaptively discarding parts of unnecessary embedding features. Our main contributions are summarized as follows:

\begin{itemize}
	\setlength\itemsep{-0.4em}
	\item A large-scale 3D high-fidelity mask face PAD dataset named HiFiMask is released. Compared with public 3D mask datasets, HiFiMask has several advantages, such as realistic masks and amount of data in the term of identities, sensors and videos.  
	\item We propose a novel Contrastive Context-aware Learning (CCL) framework to efficiently leverage rich and fine-grained context between live and mask faces for discriminative feature representation.
	\item Extensive experiments conducted on the HiFiMask and three other public 3D mask datasets demonstrate the challenges of HiFiMask and the effectiveness of the proposed method.
\end{itemize}

\section{Related Work}
\subsection{3D Mask Datasets}
Recently, several 3D mask face PAD datasets have been released. As listed in Tab.~\ref{table:datasets}, 3DMAD~\cite{nesli2013spoofing} is the first publicly available 3D mask dataset, which consists of $255$ videos from $17$ subjects, and the masks are made of paper and hard resin. Subsequent datasets 3DFS-DB~\cite{galbally2016three}, HKBU-MARs V2~\cite{liu20163d2}, and BRSU Skin/Face/Spoof (briefly named BRSU)~\cite{steiner2016reliable} improve previous drawbacks in terms of acquisition devices, mask types, and lighting environment. The recent CASIA-SURF 3DMask~\cite{yu2020nasfas} has a large number of videos under various lighting conditions using various recording sensors. Still, it has a limited number of subjects and mask types. Besides common RGB modality, several multi-modal mask datasets such as MLFP~\cite{agarwal2017face}, ERPA~\cite{bhattacharjee2017you}, and WMCA~\cite{george2019biometric} extend the study from visible light to near-infrared and thermal spectrums. 

Overall, there are three main limitations of existing 3D mask datasets: 1) a limited number of samples, resulting in potential overfitting; 2) lack of clear attribute information (\eg, skin tone and lighting) for evaluating the impact of external factors; and 3) the masks are not realistic enough in terms of color texture and structure, and they are recorded under stable lighting conditions.

\subsection{Face PAD Approaches based on 3D Mask}
Compared with 2D presentation attacks, 3D mask attacks are more realistic to live faces in terms of depth shape and color texture. There are few works to exploit fine-grained features for discrimination. Jia et al.~\cite{jia20203d} designed a two-stream network based on factorized bilinear coding of multiple color channels, targeting learning subtle, detailed cues. Yu et al.~\cite{yu2020nasfas} searched the well-suited central difference architectures with intrinsic feature representation. George et al.~\cite{george2019biometric} proposed a multi-channel CNN, which aggregated the features among RGB, depth, infrared, and thermal modalities for robust mask PAD.

On the temporal side, several rPPG-based methods~\cite{li2016generalized,liu20163d,liu2018remote,lin2019face,liu2021multi} are proposed according to the evidence that periodic rPPG pulse cues could be recovered from the live faces but noisy for the mask attacks. Li et al.~\cite{li2016generalized} was the first to leverage the facial rPPG signals' frequency statistics for mask attacks detection. Liu et al.~\cite{liu20163d,liu2018remote,liu2021multi} combined both local rPPG signals and global background noises to learn consistent rPPG features for 3D mask PAD.

As for metric learning-based PAD approaches, contrastive loss~\cite{hao2019face} and triplet loss~\cite{li2020compactnet,wang2019improving} are utilized to widen the distance between the live faces and PAs. Recently, contrastive learning~\cite{he2020momentum,chen2020simple,grill2020bootstrap,chen2020exploring} achieved outstanding performance in self-supervised generic object classification. In~\cite{khosla2020supervised}, supervised contrastive learning is proposed for boosting performance upon using basic cross-entropy loss. Despite with similar design philosophy, the proposed CCL is different from~\cite{khosla2020supervised} in both data pair generation and dropout regularization steps.

The approaches as mentioned above might be unreliable under the following situations: 1) high-fidelity mask attack with realistic appearance; 2) dynamic light flashing to disturb rPPG recovery; 3) metric learning-based constraints obtain unsatisfactory performance in PAD tasks; and 4) existing self-supervised or supervised contrastive learning approaches are not suitable for fine-grained binary classification task like 3D mask PAD. To tackle these issues, we propose a contrastive context-aware learning framework to explicitly mine the discriminative features among bonafide/mask appearance and complex scenarios.

\section{HiFiMask Dataset}
\label{sec:HiFiMask}

\begin{figure}[t]
	\centering
	\includegraphics[width=0.9\linewidth]{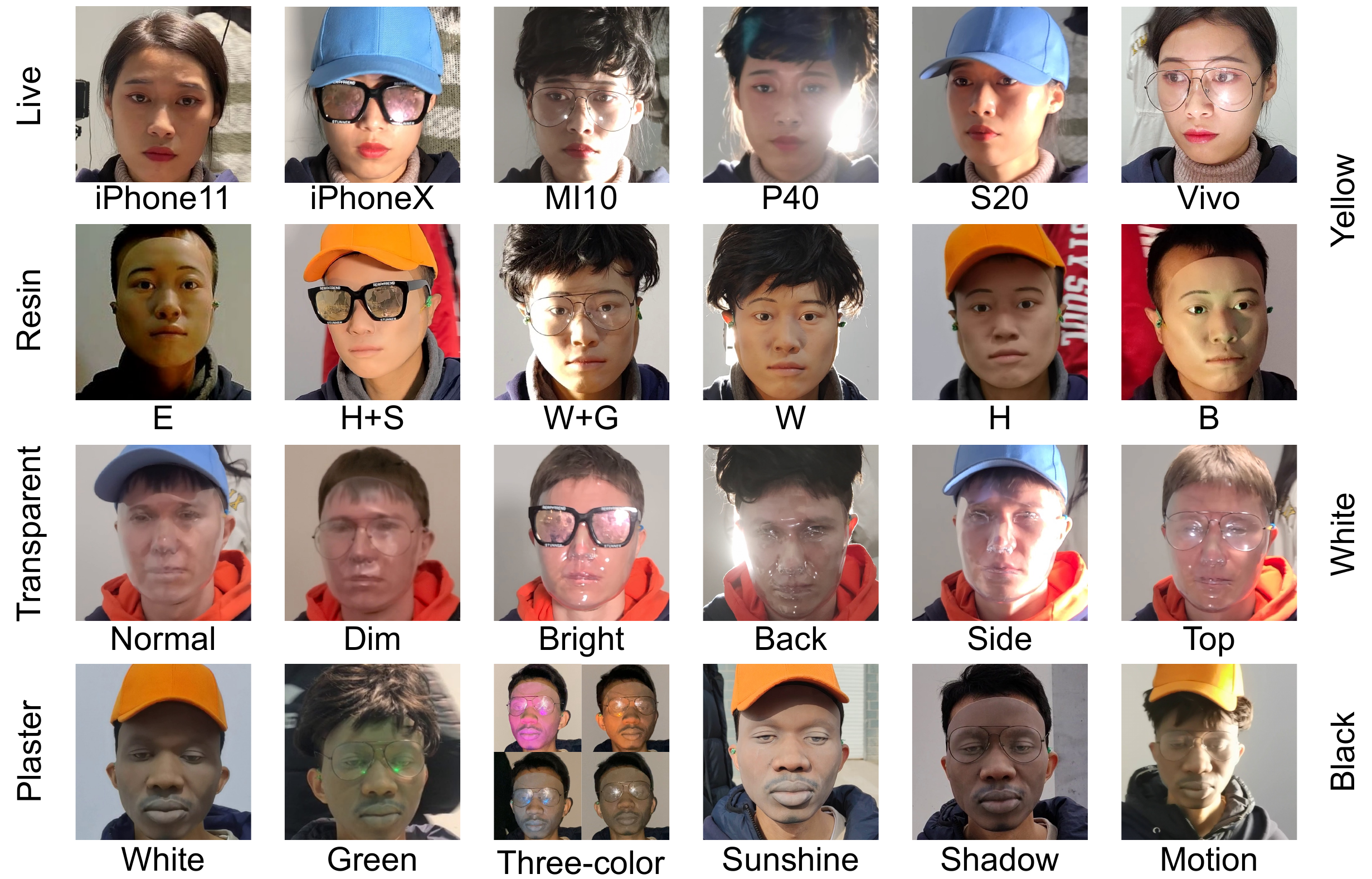}
	\caption{Samples from the HiFiMask dataset. The first row shows 6 kinds of imaging sensors. The second row shows 6 kinds of appendages, among which E, H, S, W, G, and B are the abbreviations of Empty, Hat, Sunglasses, Wig, Glasses, and messy Background, respectively. The third row shows 6 kinds of illuminations, and the fourth row represents 6 deployment scenarios.}
	\label{fig:samples}
	\vspace{-1.0em}
\end{figure}

Given the shortcomings of the current mask datasets, we carefully designed and collected a High-Fidelity Mask dataset (briefly named HiFiMask), which provides $5$ main advantages over previous existing datasets. \textbf{Advantage 1:} To the best of our knowledge, HiFiMask is currently the largest 3D face mask PAD dataset, which contains $54,600$ videos captured from $75$ subjects of three skin tones, including $25$ subjects in yellow, white, and black, respectively. \textbf{Advantage 2:} HiFiMask provides $3$ high-fidelity masks with the same identity, which are made of transparent, plaster, and resin materials, respectively. As shown in Fig.~\ref{fig:samples}, our realistic masks are visually difficult to be distinguished from live faces. \textbf{Advantage 3:} We consider $6$ complex scenes, \ie, White Light, Green Light, Periodic Three-color Light, Outdoor Sunshine, Outdoor Shadow, and Motion Blur for video recording. Among them, there is periodic lighting within [0.7, 4]Hz for the first three scenarios to mimic the human heartbeat pulse, thus might interfere with the rPPG-based mask detection technology~\cite{li2016generalized}. Please see \textsl{Appendix B} for corresponding rPPG analysis. \textbf{Advantage 4:} We repeatedly shoot $6$ videos under different lighting directions (\ie, NormalLight, DimLight, BrightLight, BackLight, SideLight, and TopLight) for each scene to explore the impact of directional lighting. \textbf{Advantage 5:} $7$ mainstream imaging devices (\ie, iPhone11, iPhoneX, MI10, P40, S20, Vivo, and HJIM) are utilized for video recording to ensure high resolution and imaging quality. Please see \textsl{\textbf{Appendix A}} for more details.

\begin{table}[]
	\centering
	\caption{Statistical information for each protocol of the proposed HiFiMask dataset. Note that 1, 2 and 3 in the fourth column mean Transparent, Plaster and Resin mask, respectively.}
	\scalebox{0.9}{
		\begin{tabular}{|c|l|c|c|c|l|l|c|l|l|c|l|l|c|l|l|}
			\hline
			\multicolumn{2}{|c|}{Pro.}                  & Subset & subject & \multicolumn{3}{c|}{Masks}   & \multicolumn{3}{c|}{\# live} & \multicolumn{3}{c|}{\# mask} & \multicolumn{3}{c|}{\# all} \\ \hline \hline
			\multicolumn{2}{|c|}{\multirow{3}{*}{1}}    & Train  & 45         & \multicolumn{3}{c|}{1\&2\&3} & \multicolumn{3}{c|}{8,108}    & \multicolumn{3}{c|}{24,406}   & \multicolumn{3}{c|}{32,514}  \\ \cline{3-16} 
			\multicolumn{2}{|c|}{}                      & Dev    & 6          & \multicolumn{3}{c|}{1\&2\&3} & \multicolumn{3}{c|}{1,084}    & \multicolumn{3}{c|}{3,263}    & \multicolumn{3}{c|}{4,347}   \\ \cline{3-16} 
			\multicolumn{2}{|c|}{}                      & Test   & 24         & \multicolumn{3}{c|}{1\&2\&3} & \multicolumn{3}{c|}{4,335}    & \multicolumn{3}{c|}{13,027}   & \multicolumn{3}{c|}{17,362}  \\ \hline
			\multicolumn{2}{|c|}{\multirow{3}{*}{2\_1}} & Train  & 45         & \multicolumn{3}{c|}{2\&3}    & \multicolumn{3}{c|}{8,108}    & \multicolumn{3}{c|}{16,315}   & \multicolumn{3}{c|}{24,423}  \\ \cline{3-16} 
			\multicolumn{2}{|c|}{}                      & Dev    & 6          & \multicolumn{3}{c|}{2\&3}    & \multicolumn{3}{c|}{1,084}    & \multicolumn{3}{c|}{2,180}    & \multicolumn{3}{c|}{3,264}   \\ \cline{3-16} 
			\multicolumn{2}{|c|}{}                      & Test   & 24         & \multicolumn{3}{c|}{1}       & \multicolumn{3}{c|}{4,335}    & \multicolumn{3}{c|}{4,326}    & \multicolumn{3}{c|}{8,661}   \\ \hline
			\multicolumn{2}{|c|}{\multirow{3}{*}{2\_2}} & Train  & 45         & \multicolumn{3}{c|}{1\&3}    & \multicolumn{3}{c|}{8,108}    & \multicolumn{3}{c|}{16,264}   & \multicolumn{3}{c|}{24,372}  \\ \cline{3-16} 
			\multicolumn{2}{|c|}{}                      & Dev    & 6          & \multicolumn{3}{c|}{1\&3}    & \multicolumn{3}{c|}{1,084}    & \multicolumn{3}{c|}{2,174}    & \multicolumn{3}{c|}{3,258}   \\ \cline{3-16} 
			\multicolumn{2}{|c|}{}                      & Test   & 24         & \multicolumn{3}{c|}{2}       & \multicolumn{3}{c|}{4,335}    & \multicolumn{3}{c|}{4,350}    & \multicolumn{3}{c|}{8,685}   \\ \hline
			\multicolumn{2}{|c|}{\multirow{3}{*}{2\_3}} & Train  & 45         & \multicolumn{3}{c|}{1\&2}    & \multicolumn{3}{c|}{8,108}    & \multicolumn{3}{c|}{16,233}   & \multicolumn{3}{c|}{24,341}  \\ \cline{3-16} 
			\multicolumn{2}{|c|}{}                      & Dev    & 6          & \multicolumn{3}{c|}{1\&2}    & \multicolumn{3}{c|}{1,084}    & \multicolumn{3}{c|}{2,172}    & \multicolumn{3}{c|}{3,256}   \\ \cline{3-16} 
			\multicolumn{2}{|c|}{}                      & Test   & 24         & \multicolumn{3}{c|}{3}       & \multicolumn{3}{c|}{4,335}    & \multicolumn{3}{c|}{4,351}    & \multicolumn{3}{c|}{8,686}   \\ \hline
		\end{tabular}
	}
	\label{tab:protocol_Statistics}
	\vspace{-0.6em}
\end{table}

\subsection{Evaluation Protocol and Statistics}
We define two protocols of HiFiMask for evaluation: Protocol 1-`seen' and Protocol 2-`unseen'. The information used in the corresponding protocol is described in Tab.~\ref{tab:protocol_Statistics}.

\noindent\textbf{Protocol 1-`seen'.}\quad
Protocol 1 is designed to evaluate algorithms' performance when the mask types have been `seen' in training and development sets. In this protocol, all skin tones, mask types, scenes, lightings, and imaging devices are presented in the training, development, and testing subsets, as shown in the second and third columns of Protocol 1 in Tab.~\ref{tab:protocol_Statistics}. 
As there are multiple liveness-unrelated factors inside, such as skin tones, lightings, and scenes, the algorithms are easily misled to learn unfaithful features, which is the challenge in this protocol.

\noindent\textbf{Protocol 2-`unseen'.}\quad
Protocol 2 evaluates the generalization performance of the algorithms for `unseen' mask types. Specifically, we further define three leave-one-type-out testing subprotocols based on Protocol 1 to evaluate the algorithm's generalization performance for transparent, plaster, and resin mask, respectively. For each protocol that is shown in the fourth columns of Protocol 2 in Tab.~\ref{tab:protocol_Statistics}, we train a model with $2$ types of masks and test on the left $1$ mask. Note that the `unseen' protocol is more challenging as the testing set's mask type is unseen in the training and development sets. Thus the learned features might be easily overfitted to the `seen' types.


\vspace{-0.5em}
\section{Methodology}
\label{sec:method}

\begin{figure*}[t]
	\centering
	\includegraphics[width=1.0\linewidth]{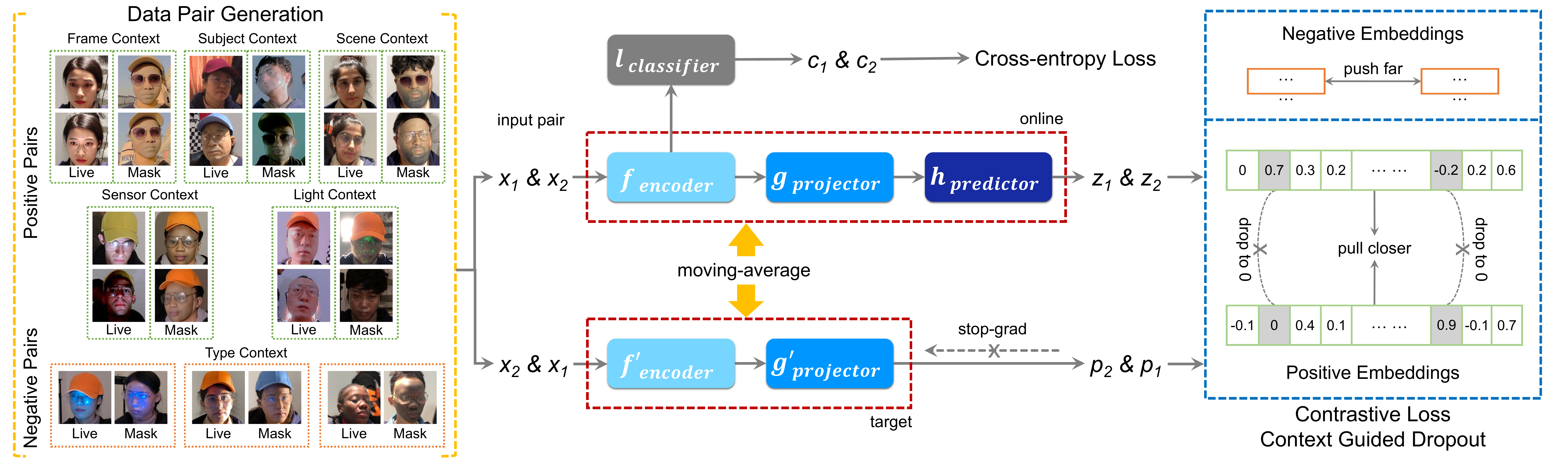}
	\caption{\textbf{CCL architecture}. The left yellow part denotes our data pair generation manner. 
		Each pair of images is processed by central framework twice, consisting of an online network (\textsl{$f$}, \textsl{$g$}, \textsl{$h$}), a target network without gradient backward (\textsl{$f'$}, \textsl{$g'$}) and a classifier header (\textsl{$l$}). The right blue part denotes the CGD module, the positive embeddings (in \textsl{$z_1$}\&\textsl{$p_2$},  \textsl{$z_2$}\&\textsl{$p_1$}) are pulled closer by CGD. \textbf{Best viewed in color.}}
	\label{fig:overall}
\end{figure*}

This section we introduces the Contrastive Context-Aware Learning (CCL) framework for 3D high-fidelity mask PAD. CCL train models by contrastive learning but in a supervised learning manner. As illustrated in Fig.~\ref{fig:overall}, CCL contains a data pair generation module to generate input data by leveraging rich contextual cues, a well-designed contrastive learning architecture for face PAD tasks, and the Context Guided Dropout (CGD) module accelerates the network convergence during the early training stages.

\subsection{Data Pair Generation}
\label{sec:data}


To effectively leverage rich contextual cues (\eg, skin, subject, type, scene, light, sensor, and inter-frame information) in HiFiMask datasets, we organize the data into pairs and freeze some of the contexts to mine the discrimination of other contexts, \eg, we select a live face and a resin mask face from the same subject. Then the contrast is the discrepancy between the material of the live skin and resin. We split the live and mask faces into fine-grained patterns to generate a variety of meaningful contextual pairs. As shown in Tab.~\ref{tab:data organize}, we generate contextual pairs in the following way:
1) in Pattern. 1, we sample two frames from one single video as one kind of positive context pair;
2) in Pattern. 5, we sample one fine-grained mask category and the living category with the same subject as one negative context pair;
3) the positive and negative context pairs, including but not limited to the above combinations, are generated as the training set.
The diagrammatic sketch of which contexts are compared in each pattern can be found in the left part of Fig.~\ref{fig:overall}. The ablation of the pattern generation is studied in Sec.~\ref{sec:ablation}.

\subsection{Network Architecture}
\label{sec:network}
Inspired by the architectures in self-supervised learning~\cite{chen2020simple,chen2020exploring,grill2020bootstrap}, we extend the self-supervised batch contrastive approach to the fully-supervised setting, allowing us to effectively leverage label information, and propose the CCL framework,  consisting of an online network and a target network for pairwise contrastive information propagation. At the same time, an extra classifier is used for explicit supervision. As shown in Fig.~\ref{fig:overall}, well-organized contextual image pairs are utilized as the inputs of the CCL. The inputs are sent to the \textsl{online} and \textsl{target} networks. The online network is composed of three modules: an encoder network \textsl{$f$}(with a backbone network and a fully connected layer), a projector \textsl{$g$} and a predictor \textsl{$h$} (with the same multi-layer perceptron structure). Similarly, the target network has an encoder \textsl{$f'$} and a projector \textsl{$g'$} with different weights from the online network. As shown in Eq.~\ref{con:moving average}, the weights of the target network $\theta'$ perceive an exponential moving-average~\cite{he2020momentum} of the online parameters $\theta$. We perform the moving-average after each step by target decay rate $\tau$ in Eq.~\ref{con:tau},
\begin{equation}
	\theta' \leftarrow \tau \theta'+(1-\tau)\theta,
	\label{con:moving average}
\end{equation}
\begin{equation}
	\tau \triangleq 1-(1-{\tau_{base}})\cdot(cos(\pi \mathit{s}/S+1)/2.
	\label{con:tau}
\end{equation} The exponential parameter $\tau_{base}$ is set to 0.996, $s$ is the current training step, and $S$ is the maximum number of training steps. In addition, a classifier head \textsl{$l$} is added after the encoder \textsl{$f$} in order to perform supervised learning.
During the inference stage, only the encoder \textsl{$f$} and classifier \textsl{$l$} are applied to perform the discrimination of mask samples.


\subsection{Context Guided Dropout}
\label{sec:cgd}
In classical self-supervised contrastive learning frameworks~\cite{chen2020simple,chen2020exploring,grill2020bootstrap}, the input images $x_1$ and $x_2$ are augmented from a source image $x$. As a result, the similarity loss between $x_1$ and $x_2$ would decrease to a relatively low level smoothly. In contrast, our proposed CCL constructs the positive contextual pairs from separate source images, which suffer from high dissimilarity, leading to unstable convergence. Moreover, the contextual features (\eg, scenes) might not always be relevant to the live/spoof cues, leading to a large similarity loss.


Inspired by the dropout operator~\cite{choe2019attention,xiao2016learning} to randomly discard parts of neurons during training, we propose Context Guided Dropout (CGD), which adaptively discards parts of the `outlier' embedding features according to their similarities. For instance, given the embeddings from positive pairs, we assume that the abnormal differences between them belong to the context information. Therefore, we could automatically drop out the abnormal embeddings with huge dissimilarities after ranking their locations. For a positive $n$-dimensional embedding pair $z_1$ and $p_2$, we first calculate the difference vector $\delta_i$ via
\begin{equation}
	\delta_i = | (\frac {z_1} {||z_1||_2})^2 - (\frac {p_2} {||p_2||_2})^2|  \label{con:diffvector}
\end{equation}
Afterward, we sort $\delta_i$ by descending sequence and record the index of the largest $p_d \cdot n$. Here $p_d$ is the proportion of embedding feature channels to be discarded. To make our method more robust, we execute this procedure in a mini-batch to determine the discarding position marked as $topK$. Besides, the discarded embedding is scaled by a factor of $1/{(1-p_d)}$, which is similar to the inversed dropout method. The complete procedure is described in Alg.~\ref{alg:CCL2}.


To make CGD more adaptive, we apply a cosine decay factor to $p_d$ during training as follows
\begin{equation}
	{p_d} \leftarrow \frac {p_d} {2} \cdot (1+cos(2 \pi \frac{q_{cur}} {q})) \cdot \mathbbm{1}_{q_{cur}<q/2} 
	\label{con:updatepd}
\end{equation}

where $q_{cur}$ is the current epoch and $q$ is the training epochs to be set.  $\mathbbm{1}_c \in \{0,1\}$ is an indicator function which returns 1 if condition $c$ is true. We also visualize the training logs in \textsl{\textbf{Appendix C}}. It can be seen that assembling CGD accelerates the network convergence during the early training stages, making the whole CCL training more stable.

\begin{algorithm}
	\small{
		\caption{Contrastive Context-Aware Learning (CCL)}
		\label{alg:CCL2}
		\begin{algorithmic}  \\
			\!\!\!\!\!\!\! \textbf{Require:} {image pair set $\mathcal{X}$ with label set $\mathcal{Y}$.}  \\
			\!\!\!\!\!\! 1: Initialize: encoder $f$, projector $g$, predictor $h$ by $\theta$ \\
			\!\!\!\!\!\! 2: Initialize: encoder $f'$, projector $g'$  by $\theta' \ (\theta'=\theta)$ \\
			\!\!\!\!\!\! 3: \textbf{while} not end of training \textbf{do} \\
			\!\!\!\!\!\! 4: \qquad \!\!\! \textbf{sample} batch $\mathbf{X}$, $\mathbf{Y}$ in $\mathcal{X}$, $\mathcal{Y}$ \qquad  // batch size=$N$  \\
			\!\!\!\!\!\! 5: \qquad \!\!\! \textbf{for} \textit{i} th ${x}_1, {x}_2, {y}_1, {y}_2$  \textbf{do}  \\
			\!\!\!\!\!\! 6: \qquad \qquad \!\!\!\!\!\! {$ {c}_1 = l \circ f({x}_1), \ {c}_2 = l \circ f({x}_2) $}   \\
			\!\!\!\!\!\! 7: \qquad \qquad \!\!\!\!\!\! {$ {z}_1 = h \circ g \circ f({x}_1), \ {z}_2 = h \circ g \circ f({x}_2) $}  \\
			\!\!\!\!\!\! 8: \qquad \qquad \!\!\!\!\!\! {$ {p}_1 = g' \circ f' ({x}_1), \ {p}_2 = g' \circ f' ({x}_2) $}  \qquad \quad \!\!\!\!\!\! // length=n \\
			\!\!\!\!\!\! 9: \qquad \qquad \!\!\!\!\!\! {\textbf{if} ${y}_1 = {y}_2$} \textbf{then} \qquad \qquad \qquad \qquad \qquad \!\! // CGD start  \\
			\!\!\!\!\!\!\!\!\! 10: \qquad \qquad \qquad \!\!\!\!\!\!\!\!\! {\textbf{compute} $\delta_i$ from ${z}_1$ and ${p}_2 $, see Eq.~\ref{con:diffvector}}   \\
			\!\!\!\!\!\!\!\!\! 11: \qquad \qquad \qquad \!\!\!\!\!\!\!\!\! {$ {k_i} = \{t|\delta_i^{(t)} > \Tilde{\delta_i}^{(p_d \cdot n)}, \Tilde{\delta_i}=sort(\delta_i),t \in [1,n] \}$ } \\
			\!\!\!\!\!\!\!\!\! 12: \qquad \qquad \qquad \!\!\!\!\!\!\!\!\! {\textbf{do} the same to ${z}_2, {p}_1 $}, get $k_i'$  \\
			\!\!\!\!\!\!\!\!\! 13: \!\!\! \qquad {$ K = [K^{(t)},...],\ K^{(t)} = \Sigma_{i=1}^{N} {(\mathbbm{1}_{t \in k_i} + \mathbbm{1}_{t \in k_i'} )},t \in [1,n] $ } \\
			\!\!\!\!\!\!\!\!\! 14: \qquad \!\!\! {$ topK = \{ t | K^{(t)} > \Tilde{K}^{(p_d \cdot n)}, \Tilde{K} = sort(K),t \in [1,n] \} $} \\
			\!\!\!\!\!\!\!\!\! 15: \qquad \!\!\! \textbf{for} {\textit{i} th $ {z}_1, {z}_2, {p}_1, {p}_2$ \ \textbf{do}} \\
			\!\!\!\!\!\!\!\!\! 16: \qquad \qquad \!\!\!\!\!\! {$ \hat{z}_1 = [\hat{z}_1^{(t)},... ]\ \textbf{if} \ t \in topK,\ \hat{z}_1^{(t)}=0; \textbf{else}\ \hat{z}_1^{(t)}=z_1^{(t)} $} \\
			\!\!\!\!\!\!\!\!\! 17: \qquad \qquad \!\!\!\!\!\! {$ \hat{z}_1 \leftarrow 1 / (1-p_d) \cdot \hat{z}_1 $} \\
			\!\!\!\!\!\!\!\!\! 18: \qquad \qquad \!\!\!\!\!\! {\textbf{do} the same to ${z}_2, {p}_1, {p}_2$, get $\hat{z}_2, \hat{p}_1, \hat{p}_2$} \quad \!\! // CGD end  \\
			\!\!\!\!\!\!\!\!\! 19: \qquad \!\!\! {\textbf{compute} $\mathcal{L}_{cls}$, $\mathcal{L}_{con}$, $\mathcal{L}_{total}$, see Eq.~\ref{con:lcls}, \ref{con:lcon}, \ref{con:fd}} \\
			\!\!\!\!\!\!\!\!\! 20: \qquad \!\!\! $\Delta \theta = backward(\mathcal{L}_{total}) $  \qquad \qquad \qquad \quad \!\!\!\!  // stop gradient \\
			\!\!\!\!\!\!\!\!\! 21: \qquad \!\!\! $\theta \leftarrow \theta - learning\_rate \cdot \Delta \theta $ \\
			\!\!\!\!\!\!\!\!\! 22: \qquad \!\!\! \textbf{update} $\theta'$ by Eq.~\ref{con:moving average}, $p_d$ by Eq.~\ref{con:updatepd}
		\end{algorithmic}
	}
\end{algorithm}

\vspace{-0.5em}
\subsection{Overall Loss}
\label{sec:loss}
As CCL supervises face PAD models with live/mask binary ground truth, it is straightforward to calculate classification loss $\mathcal{L}_{cls}$ using Binary Cross Entropy (BCE) function $f_{BCE}$, which is described in Eq.~\ref{con:lcls}. We also adopt the Eq.~\ref{con:lcon} for contrastive loss $\mathcal{L}_{con}$ calculation. To be specific, the CGD regularization $f_{cgd}$ is firstly applied to the embedding $z_1$, $p_2$ and $z_2$, $p_1$, and then Eq.~\ref{con:fd} is applied to calculate the cosine similarity between normalized $\hat z_1$ and $\hat p_2$, as well as normalized $\hat z_2$ and $\hat p_1$.  
\begin{equation}
	\mathcal{L}_{cls} = \frac {1} {2} [f_{BCE}(c_1, y_1) + f_{BCE}(c_2, y_2)]  \label{con:lcls}
\end{equation}
\begin{equation}
	\mathcal{L}_{con} = \frac {1} {2} [f_{D}(f_{cgd}(z_1, p_2)) + f_{D}(f_{cgd}(z_2, p_1))]  \label{con:lcon}
\end{equation}
\begin{equation}
	f_{D}(\hat z_1, \hat p_2) = (2 \cdot \mathbbm{1}_{y_1 \neq y_2} - 1) \cdot < {\hat z_1}, {\hat p_2} > + 1 \label{con:fd}
\end{equation}


The overall loss $\mathcal{L}_{total}$ can be calculated by the weighted summation of $\mathcal{L}_{cls}$ and $\mathcal{L}_{con}$, \ie, $\mathcal{L}_{total} = \mathcal{L}_{cls} + \lambda_{con} \cdot \mathcal{L}_{con}$, where $\lambda_{con}$ denotes a trade-off hyper-parameter. An ablation about $\lambda_{con} $ is conducted in \textsl{\textbf{Appendix C}}.



\vspace{-0.5em}
\section{Experiments}
\label{sec:Experiments}

\subsection{Experimental Settings}
\noindent\textbf{Datasets \& Protocols.}\quad 
Four datasets are used in our experiments: WMCA~\cite{george2019biometric}, CASIA-SURF 3DMask (briefly named 3DMask)~\cite{yu2020nasfas}, HKBU-MARsV2 (briefly named MARsV2)~\cite{liu20163d2}, and the proposed HiFiMask. We perform Intra Testing on HiFiMask and WMCA datasets with the `seen' and `unseen' protocols, and study Cross Testing performance on 3DMask and MARsV2 datasets when training on HiFiMask.

\noindent\textbf{Performance Metrics.}\quad
In HiFiMask and WMCA datasets, Attack Presentation Classification Error Rate (APCER), Bonafide Presentation Classification Error Rate (BPCER), and ACER~\cite{ACER} are used for performance evaluation. The ACER on the testing set is determined by the Equal Error Rate (EER) and BPCER=$1\%$ thresholds on development sets for HiFiMask and WMCA, respectively. In the Cross Testing experiments on 3DMask and MARsV2 datasets, Half Total Error Rate (HTER)~\cite{bengio2004statistical} and Area Under Curve (AUC) are adopted as evaluation metrics.


\noindent\textbf{Implementation Details.}\quad
We use a universal network ResNet50~\cite{He_2016_CVPR}, and two face PAD task networks, \eg, Aux.(Depth)~\cite{Liu2018Learning}, and CDCN~\cite{yu2020searching} with varying dimension of the last layer as backbones, and report their results as baselines. Please see \textsl{\textbf{Appendix D}} for more details, \eg, architectures and optimization.

\subsection{Ablation Study}
\label{sec:ablation}
Here we conduct ablation experiments to verify the contributions of each module of the proposed CCL on Protocol 1 of the HiFiMask dataset. 


\noindent\textbf{Effect of Architectures.}\quad
As shown in Tab.~\ref{tab:ablation1}, we select ResNet50~\cite{He_2016_CVPR} as baseline and equip it with five different contrastive-based learning strategies for comparison, such as Contrastive Loss~\cite{hao2019face}, Triplet Contrastive Loss~\cite{li2020compactnet},  SimSiam~\cite{chen2020exploring}, BYOL~\cite{grill2020bootstrap} and supervised contrastive learning method SupCon~\cite{khosla2020supervised} (SC for abbreviation). The results show that contrastive-based learning is more suitable for mining the discrepancy between live face and mask material than the vanilla ResNet50, with the ACER improvement from $4.7\%$ to $2.9\%$. 




In terms of self-supervised contrastive learning approaches, SimSiam and BYOL achieve $4.0\%$ and $2.9\%$ ACER on Protocol 1 of HiFiMask, respectively. It is clear that BYOL outperforms SimSiam by a large margin, indicating the importance of moving-average when updating network parameters. Based on the moving-average mechanism, CCL further decreases the ACER by 0.5\% compared with BYOL. This is because CCL is able to efficiently exploit the elaborated contextual pairs for fine-grained feature representation while BYOL only considers the simple augmented views. In Tab.~\ref{tab:ablation1}, we also compare CCL with the recent supervised contrastive learning method SC. The CCL achieves better performance (reducing 0.8\% ACER) than SC, which indicates the advances of our context-aware pair generation and CGD strategies for 3D mask PAD task.  

In summary, the proposed CCL extends the self-supervised contrastive approach to the supervised setting and pulls together the clusters of points that belong to the same class while simultaneously pushes apart clusters of samples from different classes in the embedding space. It allows us to effectively leverage label information to mine the differences between face skin and different kinds of mask materials. 

\noindent\textbf{Effect of Data Pair Generation.}\quad
An effective Data Pair Generation can accelerate the model convergence and guide the network to mine liveness-related features. As shown in Tab.~\ref{tab:data organize}, we take context such as identity, mask, lighting, imaging sensor, and frame information into consideration to generate data pairs with 6 different patterns. We practice these six patterns to perform experiments on protocol 1 of HiFiMask, respectively. One can observe that the data pair generated under different patterns has varied performances.  

When only negative pairs are used, it is obtained an ACER of $43.7\%$. The main reason is that no positive samples are available in the training set makes the model convergence difficult. The mixed-use of multiple positive and negative pairs can optimize the performance of the model.

\begin{table}[t]
	\centering
	\caption{The ablation results on the Protocol 1 of HiFiMask.}
	\resizebox{0.47\textwidth}{!}{
		\begin{tabular}{c|c|c|c}
			\toprule[1pt]
			Method & APCER(\%) & BPCER(\%) & ACER(\%) \\
			\hline
			ResNet50~\cite{He_2016_CVPR} &3.7 &5.7 & 4.7 \\
			ResNet50 w/ Contrastive Loss~\cite{hao2019face} &3.2 &4.2 & 3.7 \\
			ResNet50 w/ Triplet Loss~\cite{li2020compactnet} &3.5 &4.9 & 4.2 \\
			ResNet50 w/ SimSiam~\cite{chen2020exploring} &2.4 &5.7 & 4.0 \\
			ResNet50 w/ BYOL~\cite{grill2020bootstrap} &\textbf{1.7} &4.0 & 2.9 \\
			ResNet50 w/ SC~\cite{khosla2020supervised} &3.2 &3.2 &3.2 \\
			\hline
			\textbf{CCL w/o CGD} &1.9 &3.3 & 2.6 \\
			\textbf{CCL w/ reverse CGD} &2.2 &3.5 &2.8 \\
			\textbf{CCL w/ BOBE CGD} &2.0 &3.4 &2.7 \\
			\textbf{CCL} &1.8 &\textbf{3.0} & \textbf{2.4} \\
			\bottomrule[1pt]
		\end{tabular}
	}
	\label{tab:ablation1}
	\vspace{-0.5em}
\end{table}

\begin{table}[t]
	\centering
	\setlength{\tabcolsep}{5pt}
	\caption{Patterns to organize image pairs by adjusting attributes.}
	\resizebox{0.47\textwidth}{!}{
		\begin{tabular}{c|c|c|c|c|c|c|c|c}
			\toprule[1pt]
			Pat. & Subject & Type & Scene & Light & Sensor & Frame & Pair & ACER \\
			\hline
			1 & & & & & & \checkmark & pos & 3.6 \\
			2 & & & & & \checkmark & \checkmark & pos &3.9 \\
			3 & & & & \checkmark & & \checkmark & pos & 4.0 \\
			4 & & & \checkmark & & & \checkmark & pos & 3.2 \\
			5 & & \checkmark & & & & \checkmark & neg &43.7 \\
			6 & \checkmark & & & & & \checkmark & pos & 3.9 \\
			1-6 & \checkmark &\checkmark &\checkmark &\checkmark &\checkmark & \checkmark & pos\&neg & 2.4 \\
			\bottomrule[1pt]
		\end{tabular}
	}
	\label{tab:data organize}
	\vspace{-0.5em}
\end{table}

\noindent\textbf{Effect of CGD.}\quad
The proposed CGD module can accelerate the model convergence in the early stage of training and alleviate the interference of useless information to the model. As shown in Tab.~\ref{tab:ablation1}, if the CGD is removed, the performance of three indicators decreases, with an APCER, BPCER and ACER increasing from $1.8\%$, $3.0\%$, and $2.4\%$ to $1.9\%$, $3.3\%$, and $2.6\%$, respectively.

In the proposed CGD, we aim to manually discard the most dissimilar neurons in the embedding by finding their location for a positive embedding pair. In order to verify the effectiveness of removing embedding points, we introduce two variants for comparison: `reverse' CGD and `BOBE'(Break Out Both Ends) CGD. The first one removes the embedding points in the opposite way to the CGD. The second one removes the most similar and the least similar embedding points at the same time. Their performance is decreased, as expected, to $2.8\%$ and $2.7\%$ for ACER. Additional ablation study results are shown in \textsl{\textbf{Appendix C}}.

\subsection{Intra Testing}
Experiments are conducted on HiFiMask and WMCA datasets. Three backbones (\ie, ResNet50, Aux.(Depth) and CDCN) are used as baselines to report performances with (w/) or without (w/o) the proposed CCL.

\noindent\textbf{Results on HiFiMask.}\quad
As shown in Tab.~\ref{tab:HiFiMask_result}, without using the proposed CCL, the Aux.(Depth) achieves the best performance with lowest ACER (\ie, Protocol 1: $3.4\%$, Protocol 2: $11.4\%$) when compared with ResNet50 and CDCN. We integrated the three backbones into CCL framework, and the performances are improved consistently. Specifically, the ACERs of ResNet50, Aux.(Depth), CDCN are decreased from $4.7\%$, $3.4\%$, $3.6\%$  to $2.4\%$, $2.6\%$, $3.1\%$ on Protocol 1, and from $18.5\%$, $11.4\%$, $14.7\%$ to $16.7\%$, $10.7\%$, $12.7\%$ on Protocol 2, respectively.

\begin{table}[t]
	\centering
	\caption{The results of intra testing on two protocols of HiFiMask.}
	\resizebox{0.47\textwidth}{!}{
		\begin{tabular}{c|c|c|c|c}
			\toprule[1pt]
			Prot. & Method & APCER(\%) & BPCER(\%) & ACER(\%) \\
			\hline
			\multirow{7}{*}{1}
			&ResNet50~\cite{He_2016_CVPR} &3.7 &5.7 & 4.7 \\
			&Aux.(Depth)~\cite{Liu2018Learning} &4.9 &\textbf{1.8} & 3.4 \\
			&CDCN~\cite{yu2020searching} &3.3 &3.9 & 3.6 \\
			&\textbf{ResNet50 w/ CCL} &\textbf{1.8} &3.0 & \textbf{2.4} \\
			&\textbf{Aux.(Depth) w/ CCL} &2.1 &3.1 &2.6 \\
			&\textbf{CDCN w/ CCL} &3.0 &3.3 &3.1 \\
			\hline
			\multirow{6}{*}{2}
			&ResNet50~\cite{He_2016_CVPR} &22.4$\pm$15.3 &14.6$\pm$6.7 &18.5$\pm$11.0 \\
			&Aux.(Depth)~\cite{Liu2018Learning} &12.1$\pm$9.4 &11.2$\pm$9.8 & 11.4$\pm$9.0 \\
			&CDCN~\cite{yu2020searching} &12.6$\pm$7.3 &16.8$\pm$15.6 & 14.7$\pm$11.4 \\
			&\textbf{ResNet50 w/ CCL} &16.7$\pm$11.2 &16.7$\pm$12.4 & 16.7$\pm$11.2 \\
			&\textbf{Aux.(Depth) w/ CCL} &\textbf{10.7$\pm$7.5} &\textbf{10.7$\pm$9.4} & \textbf{10.7$\pm$8.4} \\
			&\textbf{CDCN w/ CCL} &13.3$\pm$10.6 &12.2$\pm$9.2 & 12.7$\pm$9.8 \\
			\bottomrule[1pt]
		\end{tabular}
	}
	\label{tab:HiFiMask_result}
	\vspace{-0.5em}
\end{table}

\begin{table*}[]
	\centering
	\caption{Comparison of the results of protocols `Seen' and `Unseen' on WMCA. The values ACER(\%) reported on testing sets are obtained with thresholds computed for BPCER=$1\%$ on development sets. `RGB-D' denotes using both RGB and depth inputs.}
	\scalebox{0.77}{
		\begin{tabular}{c|c|c|cccccccc}
			\hline
			\multirow{2}{*}{Modality} & \multirow{2}{*}{Method} & \multirow{2}{*}{Seen} & \multicolumn{8}{c}{Unseen}                                                                                                                  \\ \cline{4-11} 
			&                         &                       & Flexiblemask  & Replay        & Fakehead       & Prints         & Glasses        & Papermask     & Rigidmask      & Mean$\pm$Std            \\ \hline
			\multirow{3}{*}{RGB-D}    & MC-PixBiS~\cite{george2019deep}               & 1.8                   & 49.7          & 3.7           & 0.7            & 0.1            & 16.0           & 0.2           & 3.4            & 10.5$\pm$16.7           \\
			& MCCNN-OCCL-GMM~\cite{george2020learning}          & 3.3                   & 22.8          & 31.4          & 1.9            & 30.0           & 50.0           & 4.8           & 18.3           & 22.74$\pm$15.3          \\
			& MC-ResNetDLAS~\cite{parkin2019recognizing}           & 4.2                   & 33.3          & 38.5          & 49.6           & 3.80           & 41.0           & 47.0          & 20.6           & 33.4$\pm$14.9           \\ \hline
			\multirow{6}{*}{RGB}      & ResNet50~\cite{He_2016_CVPR}                & 40.93                 & 14.48         & 15.69         & 38.00          & 32.71          & 27.33          & 20.14         & 30.22          & 25.51$\pm$8.95          \\
			& ResNet50 w/ CCL         & 30.69                 & \textbf{4.76} & 15.37         & 24.67          & 19.03          & 16.80          & \textbf{9.51} & 17.62          & \textbf{15.39$\pm$6.51} \\
			& Aux.(Depth)~\cite{Liu2018Learning}             & 42.67                 & 13.23         & 12.52         & 47.33          & 32.18          & 23.69          & 13.92         & 40.43          & 26.19$\pm$14.13         \\
			& Aux.(Depth) w/ CCL      & 30.62                 & 7.41          & 12.76         & 40.00          & \textbf{16.11} & \textbf{10.17} & 11.67         & 27.32          & 17.92$\pm$11.66         \\
			& CDCN~\cite{yu2020searching}                    & 38.41                 & 12.10         & \textbf{8.69} & 42.67          & 30.07          & 11.67          & 11.87         & 30.38          & 21.06$\pm$13.17         \\
			& CDCN w/ CCL             & \textbf{27.14}        & 7.18          & 11.79         & \textbf{21.82} & 20.53          & 35.13          & 18.91         & \textbf{15.10} & 18.64$\pm$8.91          \\ \hline
		\end{tabular}
	}
	\vspace{-1.0em}
	\label{tab:WMCA_results}
\end{table*}

We note that the challenge of Protocol 2 lies in Protocol $2\_1$, where the transparent mask is used as the test set. The ACER of ResNet50, Aux.(Depth) and CDCN on Protocol $2\_1$ are $34.0\%$, $24.1\%$ and $30.9\%$, respectively. The reason may be the appearance of transparent mask varies from the plaster and resin masks (see Fig.~\ref{fig:samples}). 
Without bells and whistles, our method still can steadily reduce the ACER of the three benchmarks  ($32.5\%$, $22.6\%$ and $26.6\%$, respectively). 



The results from Tab.~\ref{tab:HiFiMask_result} show that the proposed CCL provides good generalization with different backbones. One could thus expect that considering newly emergent networks within the CCL framework could potentially improve current performance. 


\noindent\textbf{Results on WMCA.}\quad
The experimental results of protocols `seen' and `unseen' on WMCA are shown in Tab.~\ref{tab:WMCA_results}. The results in the third column of Tab.~\ref{tab:WMCA_results} show that all three widely used face PAD backbones (\ie, ResNet50, Aux.(Depth), and CDCN) assembled with CCL can achieve significantly lower ACER values (with a decrease of $10.24\%$, $12.05\%$, and $11.27\%$, respectively). This indicates that the proposed CCL effectively leverages the context cues (e.g., rich attack types) to learn more discriminative features. Note that all three baseline methods (\ie, MC-PixBiS~\cite{george2019deep}, MCCNN-OCCL-GMM~\cite{george2020learning}, and MC-ResNetDLAS~\cite{parkin2019recognizing}) trained on two modalities (RGB and depth)  perform well ($\textless$5\% ACER) under `seen' protocol. Our future work will apply CCL to multi-modal CNN.

In terms of the `unseen' protocol, we follow the same leave-one-type-out setting as ~\cite{george2019biometric}. From Tab.~\ref{tab:WMCA_results}, we can draw similar conclusions. The proposed CCL benefits the unseen attacks detection for all three backbones including ResNet50, Aux.(Depth) and CDCN. To be specific, compared with the vanilla ResNet50, Aux.(Depth) and CDCN, the counterparts with CCL could reduce $10.12\%$, $8.27\%$, and $2.42\%$ average ACERs, respectively. In addition, as shown in Tab.~\ref{tab:WMCA_results}, the proposed method with only RGB inputs achieves better average testing performance than MCCNN-OCCL-GMM~\cite{george2020learning} and MC-ResNetDLAS~\cite{parkin2019recognizing} with multi-modal inputs, which shows the excellent generalization capabilities of CCL.

\vspace{-0.5em}
\subsection{Cross Testing}
To further evaluate the generalization ability of CCL, we perform a cross-testing evaluation setting by training the model on the proposed HiFiMask and directly testing on the MARsV2~\cite{liu20163d2} and 3DMask~\cite{yu2020nasfas} datasets. 


As shown in Tab.~\ref{tab:cross_testing}, compared with bare backbones, the proposed  CCL obtains better performance on MARsV2 and 3DMask datasets, while Aux.(Depth) with CCL achieves the best performance under cross-testing "HiFiMask to MARsV2". For example, the HTER results are reduced $6.57\%$, $0.90\%$, and $2.33\%$, and AUC results are increased $3.79\%$, $1.44\%$, and $2.11\%$ on MARsV2 dataset for ResNet50, CDCN, and Aux.(Depth) with/without CCL. The same conclusion is obtained on the 3DMask dataset.

In Tab.~\ref{tab:cross_testing}, one can observe that the performances of Aux.(Depth) has a slight decline after equipped with the CCL in the 3DMask dataset. The main reason might be that 3DMask has only a limited attack types (plaster mask), which inhibits the advantage of our Data Pair Generation strategy to generate abundant context information.

\begin{table}[t]
	\centering	
	\caption{Cross-testing results on MARsV2 and 3DMask when training on HiFiMask.}
	\resizebox{0.47\textwidth}{!}{\begin{tabular}{c|c|c|c|c}
			\toprule[1pt]
			\multirow{2}{*}{Method} & \multicolumn{2}{c|}{HiFiMask to MARsV2}  & \multicolumn{2}{c}{HiFiMask to 3DMask} 
			\\ \cline{2-5} 		
			& HTER(\%)$\downarrow$                                   & AUC(\%)$\uparrow$                                   & HTER(\%)$\downarrow$                                   & AUC(\%)$\uparrow$                                             \\ \hline
			ResNet50~\cite{He_2016_CVPR}	              & 20.61 & 86.87 & 30.35 & 76.36    \\	
			CDCN~\cite{yu2020searching}				  & 16.56 & 90.81 & 17.28 & 89.94 	\\
			Aux.(Depth)~\cite{Liu2018Learning}						  & 9.31 & 96.31	& 16.11 & 91.32	\\
			\hline
			\textbf{ResNet50 w/ CCL}               				  &14.04 &90.66 	&25.43 &81.99	 \\
			\textbf{CDCN w/ CCL}               				  &15.66 &92.25 	&\textbf{13.97}&\textbf{93.26}	 \\
			\textbf{Aux.(Depth) w/ CCL}               				  &\textbf{6.98}&\textbf{98.42} 	&16.33 &90.67	 \\
			\bottomrule[1pt]	
			
	\end{tabular}}
	\label{tab:cross_testing}
\end{table}

\begin{figure}[t]
	\centering
	\vspace{0.5em}
	\includegraphics[width=1.0\linewidth]{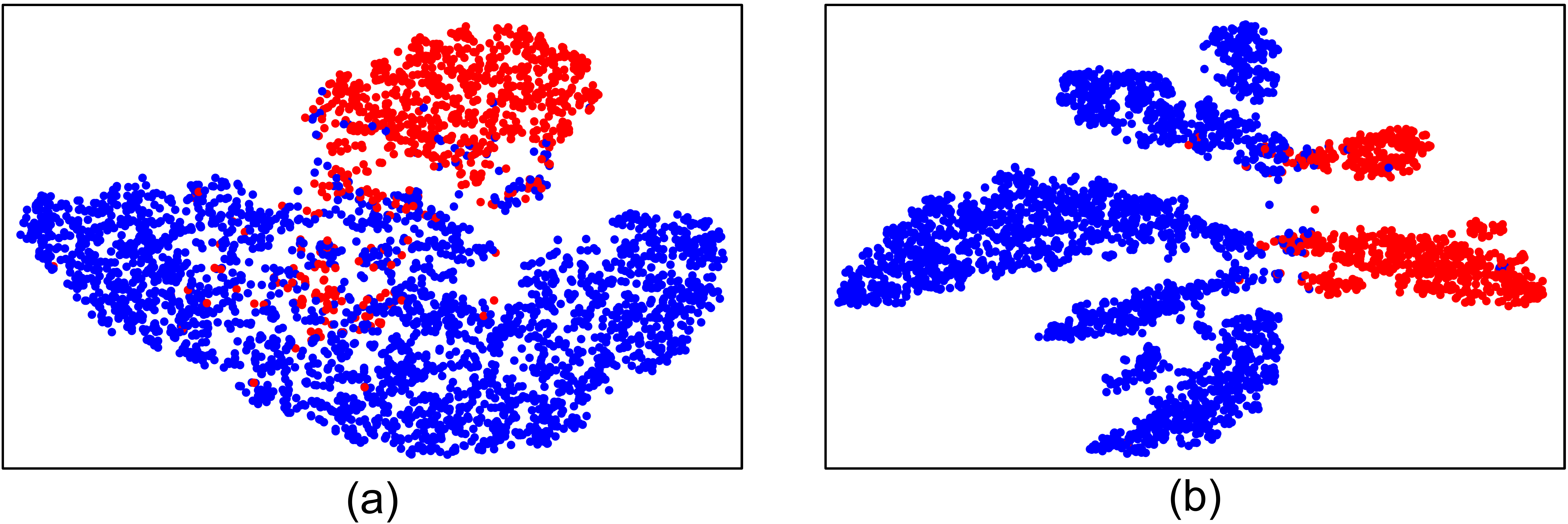}
	\vspace{-1.0em}
	\caption{Feature distribution comparison on Protocol 1 of HiFiMask using t-SNE~\cite{maaten2008visualizing}: (a) ResNet50, (b) the proposed CCL. The points with different colors denote features from different classes (red: real faces; blue: mask samples). \textbf{Best viewed in color.}}
	\label{fig:tSNE}
\end{figure}

\begin{figure}[t]
	\centering
	\vspace{0.5em}
	\includegraphics[width=1.0\linewidth]{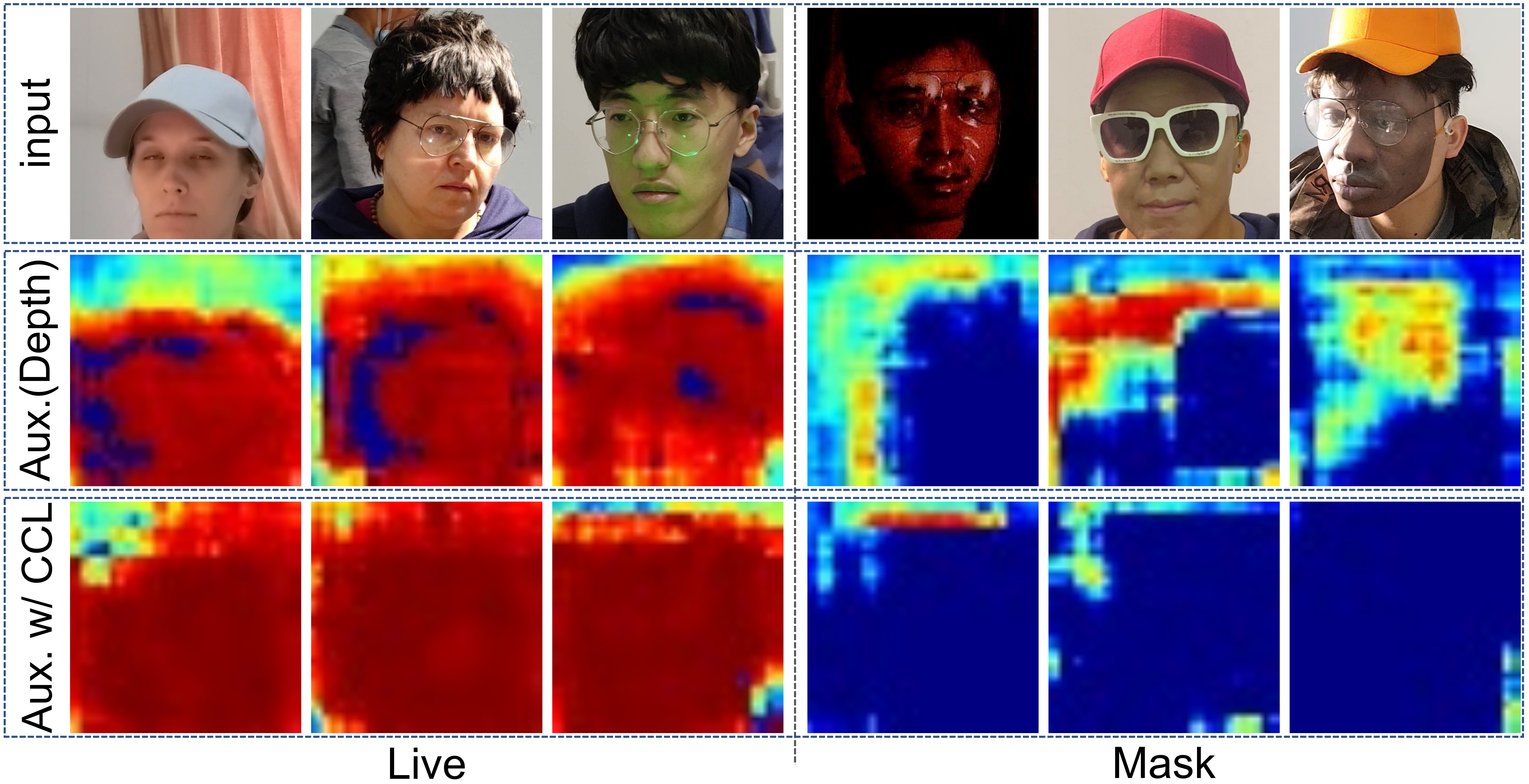}
	\vspace{-1.0em}
	\caption{ Visualization of several samples in HiFiMask. The maps are performed from Aux.(Depth) in the second row and Aux. w/ CCL in the third row (Red color: the higher the better for live samples; Blue color: the higher the better for mask samples.). \textbf{Best viewed in color.}}
	\label{fig:heatmap}
\end{figure}

\subsection{Analysis and Visualization.}
In this section, we further visually analyze the CCL's ability to distinguish different material features. 
As shown in Fig.~\ref{fig:tSNE}, we compare the features learned by ResNet50 and CCL on HiFiMask (Protocol 1).
Compared with ResNet50, the proposed CCL can well distinguish real faces from mask samples. In addition, we visualize the regression results (heatmaps) by  Aux.(Depth) and Aux.(Depth) with CCL in Fig.~\ref{fig:heatmap}. We can see that Aux.(Depth) fails to distinguish the complex mask samples in the last two columns while the proposed Aux.(Depth) with CCL provides more correct predictions. It demonstrates the better discriminative representation capacity of the proposed CCL.


\vspace{-0.5em}
\section{Conclusion}
In this paper, we release a large-scale 3D High-Fidelity Mask dataset, HiFiMask, with two challenging protocols. We hope it would push the cutting-edge research in 3D Mask Face PAD further. Besides, we propose a novel Contrastive Context-aware Learning (CCL) framework to learn discriminability by leveraging rich contexts among pairs of live faces and high-fidelity mask attacks. At last, comprehensive experiments are conducted on both HiFiMask and the other three 3D mask datasets to verify the significance of the proposed dataset and method.  

{
\bibliographystyle{ieee_fullname}
\bibliography{egbib}
}

\clearpage

\section*{Appendix}

\section*{A. Acquisition Details of HiFiMask}
Here we review the HiFiMask acquisition details in terms of equipment preparation, collection rules, and data pre-processing.

\noindent\textbf{Equipment Preparation.}\quad In order to avoid identity information to interfere with the algorithm design, the plaster, transparent and resin masks are customized for real people. We use pulse oximeter CMS60C to record real-time Blood Volume Pulse (BVP) signals and instantaneous heart rate data from live videos. For scenes of White light, Green light, Periodic Three-color light (Red, Green, Blue and their various combinations), we use a colorful lighting to set the periodic frequency of illumination changes which is consistent with the rang of human heart rate. The change frequency is randomly set between [0.7,4]Hz and recorded for future research. At the same time, we use an additional light source to supplement the light from 6 directions (NormalLight, DimLight, BrightLight, BackLight, SideLight, and TopLight). The light intensity is randomly set between 400-1600 lux.

\noindent\textbf{Collection Rules.}\quad  To improve the video quality, we pay attention to the following steps during the acquisition process: 1) All masks are worn on the face of a real person and uniformly dressed to avoid the algorithm looking for clues outside the head area; 2) Collectors were asked to sit in front of the acquisition system and look towards the sensors with small head movements; 
3) During data collection stage, a pedestrian was arranged to walk around in the background to interfere with the algorithm to compensate the reflected light clues from the background~\cite{liu2018remote};
4) All live faces or masks were randomly equipped with decorations, such as sunglasses, wigs, ordinary glasses, hats of different colors, to simulate users in a real environment.

\noindent\textbf{Data Pre-processing.}\quad
In order to save storage space, we remove irrelevant background areas from original videos, such as the part below the neck. As shown in Fig.~\ref{fig:sup_samoles}, the reserved face area is obtained through the following steps. For each video, we first use Dlib~\cite{dlib09} to detect the face in each frame and save its coordinates. Then find the largest box from all the frames of in videos to crop the face area. After face detection, we sample 10 frames at equal intervals from each video.
Finally, we name the folder of this video according with the following rule: $Skin\_Subject\_Type\_Scene\_Light\_Sensor$. Note that for the rPPG baseline~\cite{li2016generalized}, we use the first 10-second frames of each video for rPPG signal recovery without frame downsampling.

In Fig.~\ref{fig:sup_samoles}, we show some samples of one subject with yellow skin tone. Six modules with different background lighting colors represent $6$ kinds of scenes including white, green, three-color, sunshine, shadow and motion. The top of each module is the sample label or mask type, and the bottom is the scene type. Each row in one module corresponds to $7$ types of imaging sensors (one frame is randomly selected for each video), and each column shows $6$ kinds of lights.

\begin{figure}
	\vspace{-0.6em}
	\centering
	\includegraphics[scale=0.28]{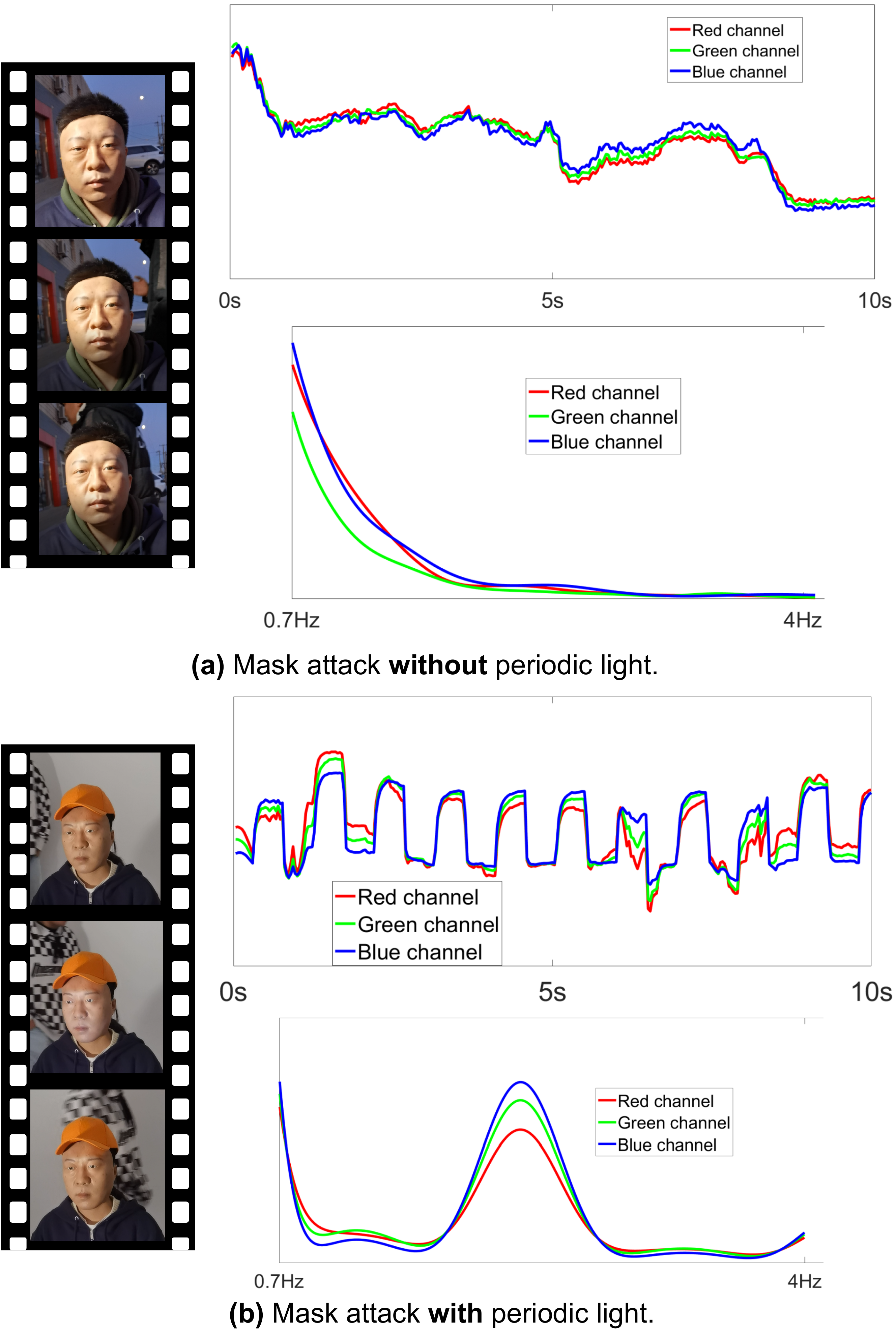}
	\caption{\small{
			Visualization of video samples and their extracted rPPG signal and power spectral density (PSD). In both cases, i.e., without (a) and with (b) external periodic light, the top right sub-figures illustrate the rPPG signal in time domain while the bottom right ones show their PSD in frequency domain. It can be seen from (b) that with period light, mask attacks could also contain pseudo `live' pulse clues. \textbf{Best viewed in color}.}
	}
	
	\label{fig:rPPG}
	\vspace{-0.6em}
\end{figure}

\section*{B. rPPG Signal Recovery from 3D Mask Attacks w/ and w/o Periodic Lights}
Here we show a simple example to illustrate the challenges on the proposed HiFiMask about the rPPG recovery. We follow the approach GrPPG~\cite{li2016generalized} to firstly track the facial region of interests (ROI), and then the intensity values within the ROI from each color channel are averaged to form the rPPG signals. As illustrated in the top right sub-figure of Fig.~\ref{fig:rPPG}(a), the rPPG signals extracted from mask attack without periodic light are quite noisy, indicating the weak heartbeat-derived liveness clues. In contrast, it can be seen from Fig.~\ref{fig:rPPG}(b) that, under the scenario with periodic light within [0.7,4]Hz, the recovered rPPG signals are with rich periodicity. We can also see from the bottom right sub-figures of Fig.~\ref{fig:rPPG}(a)(b) that the power spectral density (PSD) distributions of the extracted rPPG signals are clear to show periodicity/liveness evidences. Thus, it would mislead the rPPG-based attack detector for incorrect decision, i.e., treating the mask face with periodic light as a live face.    

In the HiFiMask dataset, as the temporal light conditions are diverse for both the bonafides and 3D mask attacks, it is not easily to use current rPPG approaches for robust PAD. Moreover, the dynamic background with pedestrian movement makes it more challenging for global noises compensation used in recent rPPG methods~\cite{liu2018remote,liu2021multi}. As there are finger-contacted BVP signals as well as instantaneous heart rate values as groundtruth for the bonafides, in the future one possible direction is to design a robust rPPG extractor and liveness detector even under complex temporal light interference. 

\section*{C. Ablation Study on Parameter $\lambda_{con}$ and CGD}
\begin{figure}[t]
	\centering
	\includegraphics[width=1.0\linewidth]{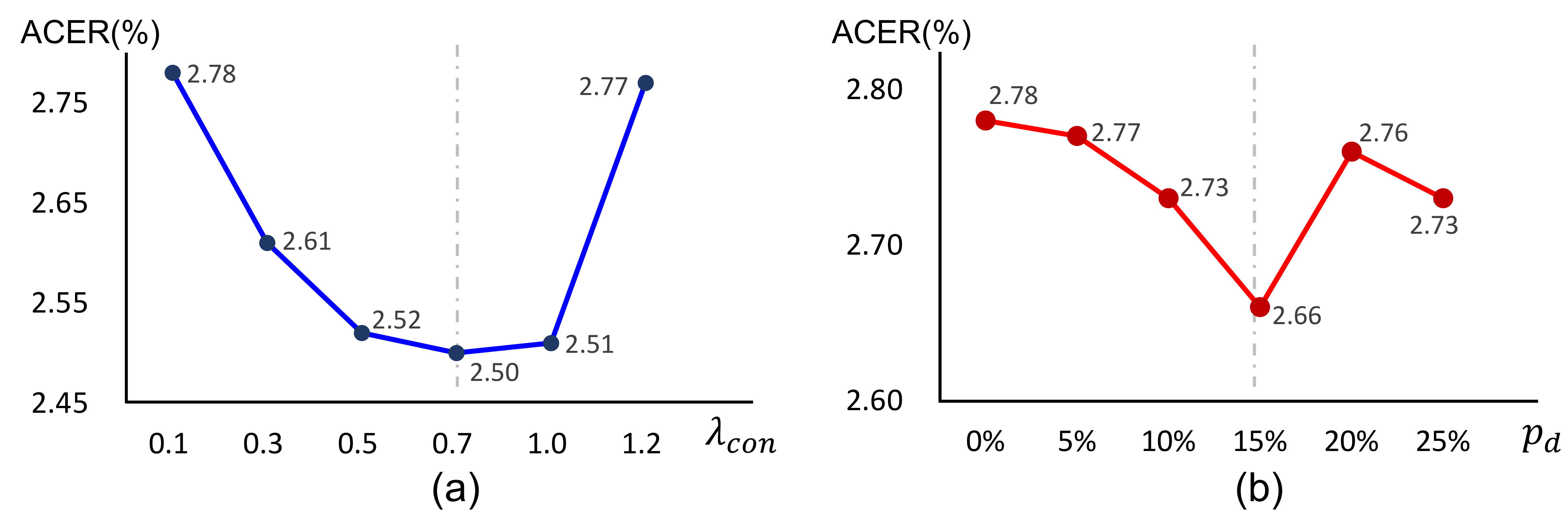}
	\caption{Ablation studies on CCL(a) and CGD(b) on the Protocol 1 of HiFiMask.}
	\label{fig:ablation results}
\end{figure}

Derived from the popular dropout operator, we can foresee that the effect of CGD would be affected by the probability of randomly abandoning neurons during training time. As shown in Fig.~\ref{fig:ablation results} (b), when the probability $p_d$ increases from $0$ to $15\%$, the performance increases from $2.78$ to $2.66$ for ACER. However, as the $p_d$ continues to increase, some functional neurons are lost, leading to performance degradation. Similarly, factor $\lambda_{con}$ in Sec.~{\color{red} 4.4} controls the relative importance of the CCL loss $\mathcal{L}_{con}$ and BCE loss $\mathcal{L}_{cls}$ in overall loss $\mathcal{L}_{tolal}$. See from the Fig.~\ref{fig:ablation results} (a), when the proportion of $\mathcal{L}_{con}$ is $0.7$, the model reaches the optimal performance in Protocol 1 of HiFiMask, that is, the ACER reaches the minimum value of $2.50\%$. Based on the above experiments, we fixed the probability $p_d$ at $15\%$ and $\lambda_{con}$ at $0.7$ in subsequent experiments.

\begin{figure}[t]
	\centering
	\includegraphics[width=1.0\linewidth]{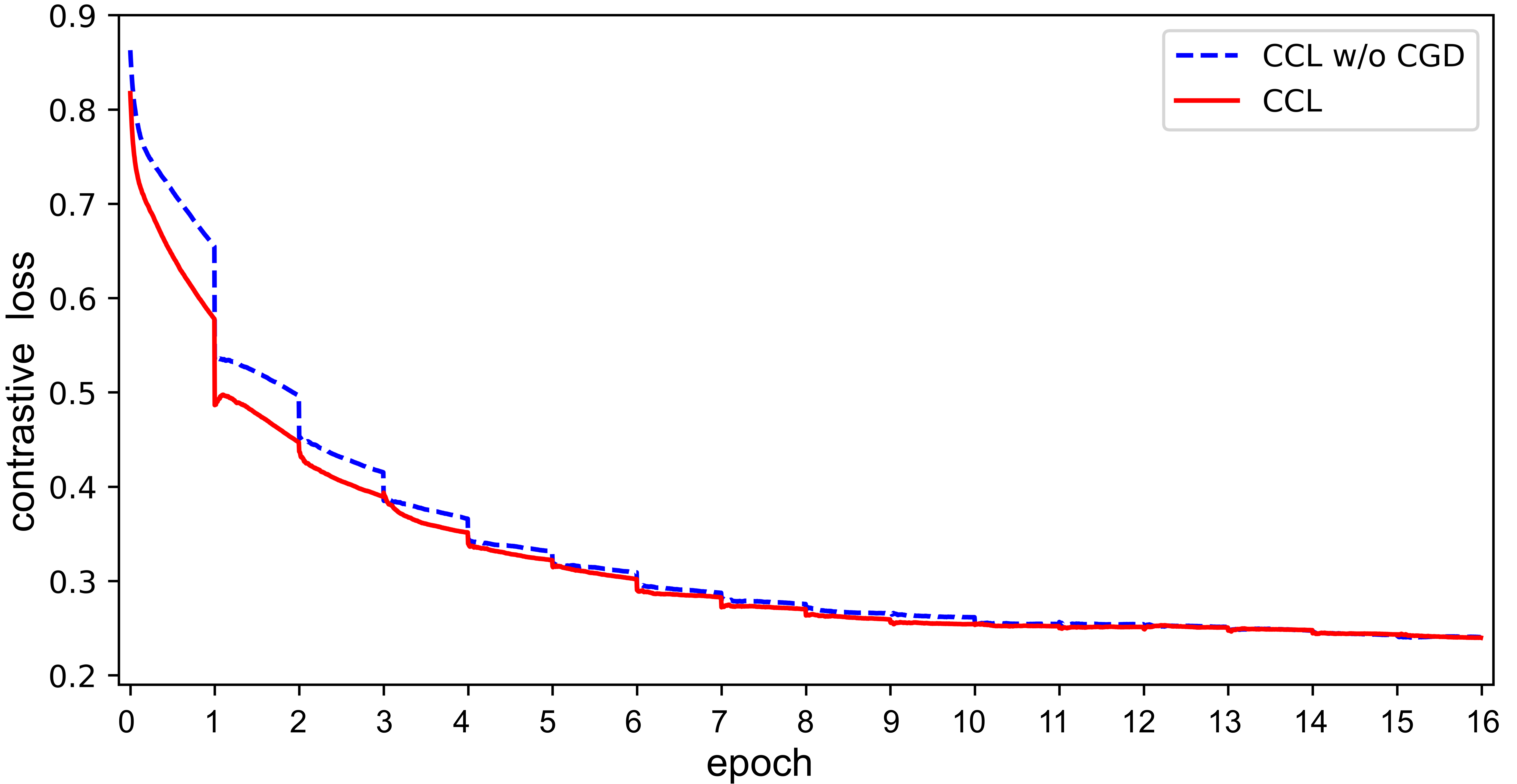}
	\caption{The contrastive loss during training stage. (Red line: CCL with CGD; Blue line: CCL without CGD.) \textbf{Best viewed in color.}}
	\label{fig:loss curve}
\end{figure}

To study the impact of the Context Guided Dropout (CGD) during the training stage, we also draw the loss curve shown in Fig.~\ref{fig:loss curve}. Intuitively, It can be seen that the red line (with CGD) drops faster than the blue line (without CGD) in the early training stage, which proves that CGD accelerates the network convergence during the early training stages making the whole CCL training more stable.

\section*{D. The Details of Architecture and Optimization.}
\noindent\textbf{Architecture.}\quad 
The last full connected(FC) layer in encoder \textsl{$f$} and \textsl{$f'$} is set from $2048$ dimensions to $128$ dimensions for ResNet50 backbone. As in SimCLR~\cite{chen2020simple}, a projector \textsl{$g$} and \textsl{$g'$} is introduced behind encoder. The projector consists of a hidden FC with $512$ dimension output followed by batch normalization~\cite{ioffe2015batch} and ReLU layers. The last layer in projector is FC only with output dimension $128$. After the projector, predictor \textsl{$h$} has the same architecture as projector.

\noindent\textbf{Optimization.}\quad 
We adopt SGD with weight decay $0.0001$ and momentum $0.9$ for the model training. The total batch size is $256$ on eight 2080Ti GPUs. The learning rate starts with $0.01$, and decays by $\gamma=0.2$ once the number of epoch reaches one of the milestones. Unless specified, models are trained for $30$ epochs with milestones in $\textsl{15, 21, 26}$.

\section*{E. ROC Results on HiFiMask.}
\begin{figure}[t]
	\centering
	\includegraphics[width=1.0\linewidth]{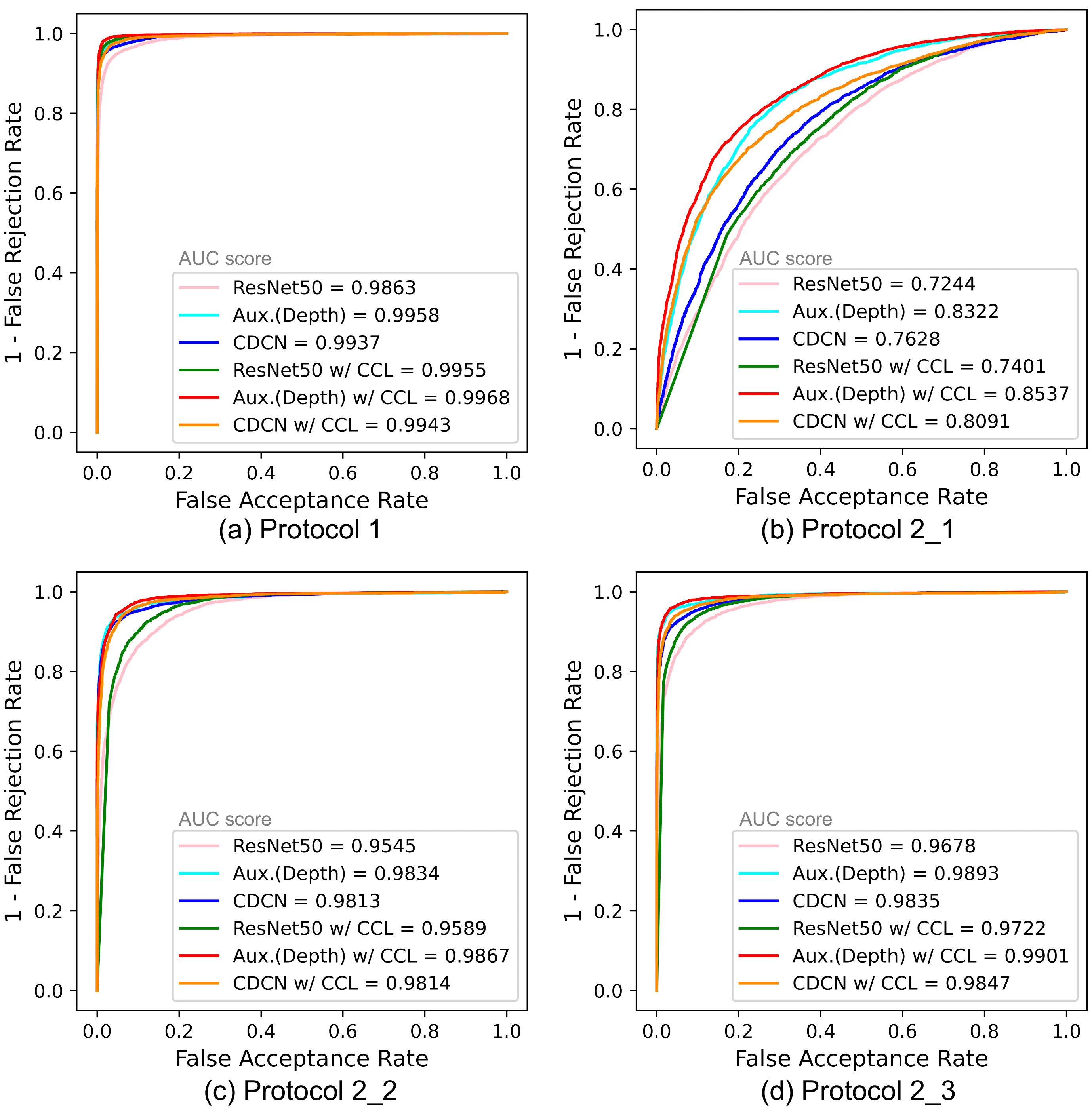}
	\caption{The ROC intra testing on two protocols of HiFiMask. \textbf{Best viewed in color.}}
	\label{fig:roc}
\end{figure}

\begin{figure*}[t]
	\centering
	\includegraphics[width=0.90\linewidth]{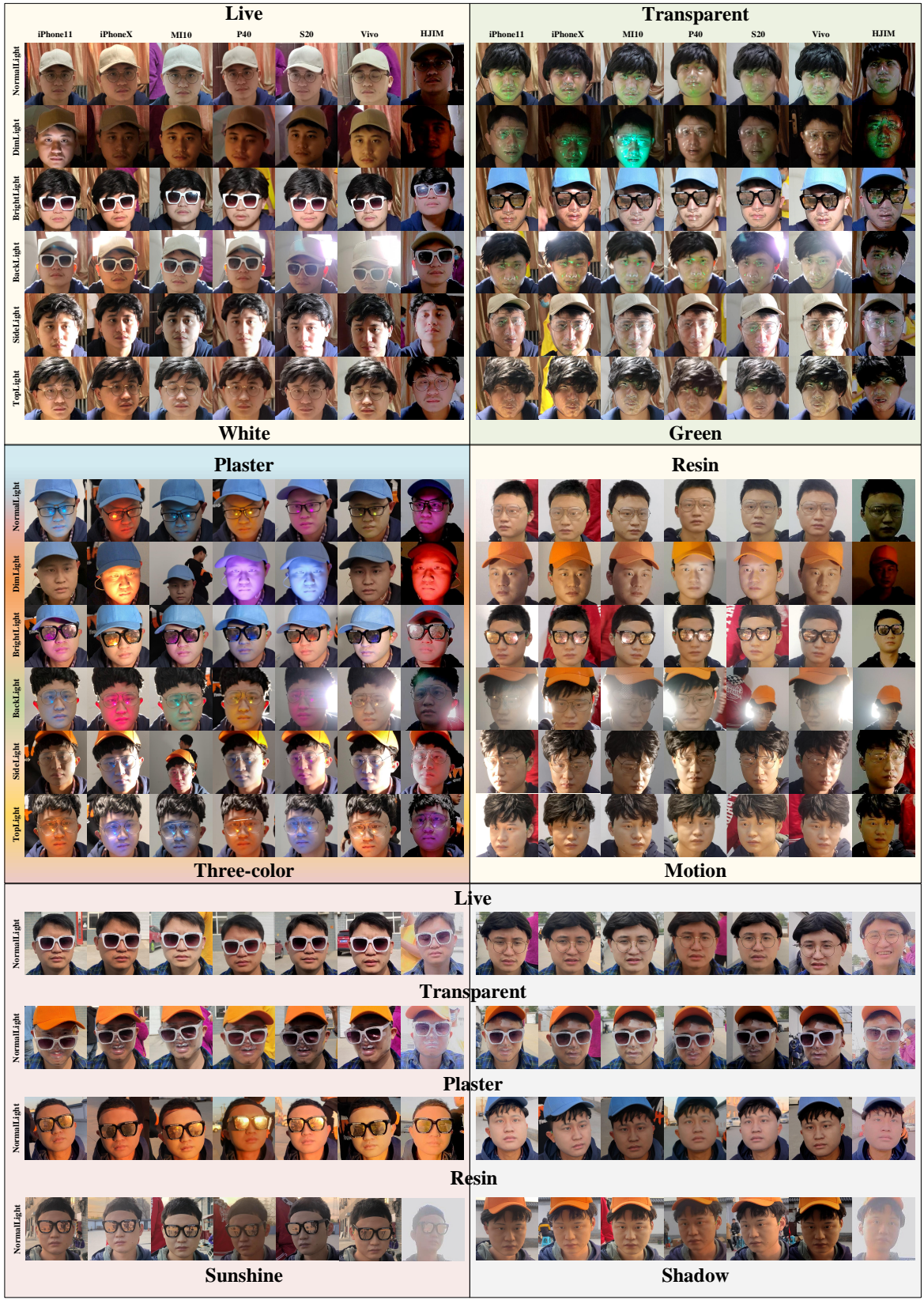}
	\caption{Samples from one subject. \textbf{Best viewed in color}.}
	\label{fig:sup_samoles}
\end{figure*}

\noindent\textbf{Protocol 1.}\quad 
To better understand the result of our proposed CCL in Tab.~{\color{red} 5}, the Receiver Operating Characteristic (ROC) curve~\cite{bi2003regression} is plotted. As shown in Fig.~\ref{fig:roc}(a), AUC (Area Under the Curve) score is calculated to measure each model's discriminative ability. AUC scores of the three baseline architectures (ResNet50, Aux(Depth), CDCN) are 0.9863, 0.9958 and 0.9937, respectively. After equipping these architectures with CCL, the AUC scores slightly increase to 0.9955, 0.9968 and 0.9943.

\noindent\textbf{Protocol 2.}\quad 
In this part, the ROC curve on the three sub-protocols of Protocol 2 are drawn in Fig.~\ref{fig:roc}(b), (c), (d) in order. Among the three sub-protocols, there is a noteworthy increment between the basic architectures (ResNet50, Aux (Depth), CDCN) and architectures with CCL. To be noted the increment in Protocol 2\_1 is more noticeable than the other two sub-protocols, which means that CCL is especially useful for a hard discriminable sub-protocol in the proposed HiFiMask.

\section*{F. More Cross Testings on 3DMask and MARsV2.}
In order to complement Sec.~{\color{red} 5.4}, we conduct another two cross-testing experiments between 3DMask and MARsV2 datasets. As shown in Tab.~\ref{tab:cross_testing2}, first we train a model on 3DMask dataset and directly test on MARsV2 dataset, then repeat the procedure by exchanging the two datasets.


Comparing the results in Tab.~{\color{red} 7} and Tab.~\ref{tab:cross_testing2}, it is obvious to find that deep models generalize better to both 3DMask and MARsV2 datasets when trained on our proposed HiFiMask. For instance, three models (ResNet50, CDCN, and Aux(Depth)) achieve 20.61\%, 16.56\% and 9.31\% HTER values when trained on HiFiMask and tested on MARsV2, respectively. These values increase to 37.96\%, 45.20\% and 44.24\% when the training set changes to 3DMask. Furthermore, there are significant drops of AUC values from 86.87\%, 90.81\% and 96.31\% to 67.05\%, 56.13\% and 57.05\%, respectively. This phenomenon can be observed as well when the testing set is 3DMask dataset. From this, we can conclude that our proposed HiFiMask is a well-distributed mask dataset, which is more helpful to train models with better generalization capacity than MARsV2 and 3DMask.

\begin{table}[t]
	\centering	
	\caption{Cross-testing results on MARsV2 and 3DMask.}
	\resizebox{0.47\textwidth}{!}{\begin{tabular}{c|c|c|c|c}
			\hline
			\multirow{2}{*}{Method} & \multicolumn{2}{c|}{3DMask to MARsV2}  & \multicolumn{2}{c}{MARsV2 to 3DMask} 
			\\ \cline{2-5} 		
			& HTER(\%)$\downarrow$                                   & AUC(\%)$\uparrow$                                   & HTER(\%)$\downarrow$                                   & AUC(\%)$\uparrow$                                             \\ \hline
			ResNet50~\cite{He_2016_CVPR}	              &37.96 &67.05 &43.69 &59.03    \\	
			CDCN~\cite{yu2020searching}				  &45.20 &56.13 & 32.56 &73.60 	\\
			Aux.(Depth)~\cite{Liu2018Learning}						  &44.24 &57.05	&43.19 &60.08	\\
			\hline
	\end{tabular}}
	\label{tab:cross_testing2}
\end{table}

\section*{G. More Study on Data Pair Increment.}
\vspace{-0.4em}
As illustrated in Tab.~\ref{tab:pair addding}, we performed such ablation experiments by combining different kinds of image pairs. After adding Sensor, Light, Scene, Type and Subject image pairs sequentially, our training set becomes larger with an increasing numbers of contrastive categories. As a result, the ACER is decreased from 3.9\% to 2.4\%, which shows a significant effect of the proposed contrastive patterns in our experiment. Note that Frame is always kept as a considerable variable throughout our ablation study.

\begin{table}[t]
	\centering
	\setlength{\tabcolsep}{5pt}
	\caption{Ablation study of combining different image pairs.}
	\resizebox{0.47\textwidth}{!}{
		\begin{tabular}{c|c|c|c|c|c|c|c|c|}
			\toprule[1pt]
			Pat. & Subject & Type & Scene & Light & Sensor & Frame & Pair & ACER(\%) \\
			\hline
			1 & & & & & \checkmark & \checkmark & pos &3.9 \\
			2 & & & & \checkmark & \checkmark & \checkmark & pos & 3.2 \\
			3 & & & \checkmark & \checkmark & \checkmark & \checkmark & pos & 3.0 \\
			4 & & \checkmark & \checkmark & \checkmark & \checkmark & \checkmark & neg & 2.9 \\
			5 & \checkmark &\checkmark &\checkmark &\checkmark &\checkmark & \checkmark & pos\&neg & 2.4 \\
			\bottomrule[1pt]
		\end{tabular}
	}
	\label{tab:pair addding}
	\vspace{-0.5em}
\end{table}

\begin{table}[]
	\centering
	\caption{Statistical information for protocol 3 of the proposed HiFiMask dataset. }
	\scalebox{0.85}{
		\begin{tabular}{|c|l|c|c|c|l|l|c|l|l|c|l|l|c|l|l|}
			\hline
			\multicolumn{2}{|c|}{Pro.}                  & Subset & subject & \multicolumn{3}{c|}{Masks}   & \multicolumn{3}{c|}{\# live} & \multicolumn{3}{c|}{\# mask} & \multicolumn{3}{c|}{\# all} \\  \hline
			\multicolumn{2}{|c|}{\multirow{3}{*}{3}}    & Train  & 45         & \multicolumn{3}{c|}{1\&3} & \multicolumn{3}{c|}{1,610}    & \multicolumn{3}{c|}{2,105}   & \multicolumn{3}{c|}{3,715}  \\ \cline{3-16} 
			\multicolumn{2}{|c|}{}                      & Dev    & 6          & \multicolumn{3}{c|}{1\&3} & \multicolumn{3}{c|}{210}    & \multicolumn{3}{c|}{320}    & \multicolumn{3}{c|}{536}   \\ \cline{3-16} 
			\multicolumn{2}{|c|}{}                      & Test   & 24         & \multicolumn{3}{c|}{1\&2\&3} & \multicolumn{3}{c|}{4,335}    & \multicolumn{3}{c|}{13,027}   & \multicolumn{3}{c|}{17,362}  \\ 
			\hline
		\end{tabular}
	}
	\label{tab:protocol-3}
	\vspace{-0.6em}
\end{table}


\vspace{-0.5em}
\section*{H. Protocol 3 of HiFiMask.}
\vspace{-0.5em}
\noindent\textbf{Protocol 3-`open set'.}\quad
Protocol 3 evaluates both discrimination and generalization ability of the algorithm under the open set scenarios. In other words, the training and developing sets contain only parts of common mask types and scenarios while there are more general mask types and scenarios on testing set. As shown in Tab.~\ref{tab:protocol-3}, based on Protocol 1, we define training and development sets with parts of representative samples while full testing set is used. Thus, the distribution of testing set is more complicated than the training and development sets in terms of mask types, scenes, lighting, and imaging devices. Different from Protocol 2 with only ‘unseen’ mask types, Protocol 3 considers both `seen' and ‘unseen’ domains as well as mask types, which is more general and valuable for real-world deployment.

\noindent\textbf{Results on Protocol 3.}\quad
As shown in Tab.~\ref{tab:protocol-3 results}, we conduct a baseline test on Protocol 3. The proposed CCL also improves the performance of all three models (ResNet50, CDCN, and Aux.(Depth)) with a convincing margin on the open set testing. It is interesting to find that the ACERs on Protocol 3 are far worse than those on Protocol 1 in Tab.~\ref{tab:HiFiMask_result}, indicating the challenges when training on only parts of common mask types and scenarios.

\begin{table}[t]
	\centering	
	\caption{The results of intra testing on Protocol 3 of HiFiMask.}
	\resizebox{0.43\textwidth}{!}{\begin{tabular}{c|c|c|c}
			\hline
			Method & APCER(\%) & BPCER(\%) & ACER(\%)  \\ 
			\hline
			ResNet50~\cite{He_2016_CVPR} & 13.5 & 28.3 & 20.9    \\	
			CDCN~\cite{yu2020searching} & 20.8 & 12.5 & 16.7 	 		\\
			Aux.(Depth)~\cite{Liu2018Learning} & 9.6 & 16.2 & 12.9	\\
			\hline
			ResNet50 w/ CCL  & 13.8 & 23.4 & 18.6 \\
			CDCN w/ CCL  & 15.4 & 13.2 & 14.3 \\
			Aux.(Depth) w/ CCL  & \textbf{8.2} & \textbf{12.7} & \textbf{10.5} \\
			\hline
	\end{tabular}}
	\vspace{-1.2em}
	\label{tab:protocol-3 results}
\end{table}

\end{document}